\documentclass{article}

\usepackage{microtype}
\usepackage{graphicx}
\usepackage{multirow}
\usepackage{mathtools}
\usepackage{subcaption}
\usepackage{booktabs} 

\PassOptionsToPackage{hyphens}{url}\usepackage{hyperref}

\usepackage[accepted]{icml2023}

\usepackage{amsmath}
\usepackage{amssymb}
\usepackage{mathtools}
\usepackage{amsthm}

\usepackage[capitalize,noabbrev]{cleveref}

\theoremstyle{plain}

\theoremstyle{definition}

\theoremstyle{remark}

\usepackage[textsize=tiny]{todonotes}

\begin{document}

\twocolumn[
\icmltitle{Future-conditioned Unsupervised Pretraining for Decision Transformer}




\icmlsetsymbol{equal}{*}

\begin{icmlauthorlist}
\icmlauthor{Zhihui Xie}{sjtu}
\icmlauthor{Zichuan Lin}{tencent}
\icmlauthor{Deheng Ye}{tencent}
\icmlauthor{Qiang Fu}{tencent}
\icmlauthor{Wei Yang}{tencent}
\icmlauthor{Shuai Li}{sjtu}

\end{icmlauthorlist}

\icmlaffiliation{sjtu}{John Hopcroft Center for Computer Science, Shanghai Jiao Tong University, Shanghai, China}

\icmlaffiliation{tencent}{Tencent AI Lab, Shenzhen, China}

\icmlcorrespondingauthor{Deheng Ye}{dericye@tencent.com}
\icmlcorrespondingauthor{Shuai Li}{shuaili8@sjtu.edu.cn}

\icmlkeywords{Machine Learning, Reinforcement Learning, Unsupervised Pretraining, ICML}

\vskip 0.3in
]



\printAffiliationsAndNotice{Work
done during the first author’s internship at Tencent AI Lab.}  

\makeatletter
\def\thickhline{%
  \noalign{\ifnum0=`}\fi\hrule \@height \thickarrayrulewidth \futurelet
   \reserved@a\@xthickhline}
\def\@xthickhline{\ifx\reserved@a\thickhline
               \vskip\doublerulesep
               \vskip-\thickarrayrulewidth
             \fi
      \ifnum0=`{\fi}}
\makeatother

\newlength{\thickarrayrulewidth}
\setlength{\thickarrayrulewidth}{2\arrayrulewidth}
\newcommand{\zhihui}[1]{{\color{green}[#1]}}

\newcommand{\statee}{\boldsymbol{s}}
\newcommand{\action}{\boldsymbol{a}}
\newcommand{\future}{\boldsymbol{z}}

\newcommand{\ours}{PDT}

\definecolor{mypurple}{RGB}{127, 0, 127}
\definecolor{myblue}{RGB}{157, 195, 230}
\definecolor{mygreen}{RGB}{112, 173, 71}
\definecolor{myred}{RGB}{244, 177, 131}
\definecolor{mygray}{RGB}{146, 146, 146}
\begin{abstract}
Recent research in offline reinforcement learning (RL) has demonstrated that return-conditioned supervised learning is a powerful paradigm for decision-making problems.
While promising, return conditioning is limited to training data labeled with rewards and therefore faces challenges in learning from unsupervised data.
In this work, we aim to utilize generalized future conditioning to enable efficient unsupervised pretraining 
from reward-free and sub-optimal offline data.
We propose Pretrained Decision Transformer ({\ours}), a conceptually simple approach for unsupervised RL pretraining.
{\ours} leverages future trajectory information as a privileged context to predict actions during training.
The ability to make decisions based on both present and future factors enhances {\ours}'s capability for generalization.
Besides, this feature can be easily incorporated into a return-conditioned framework for online finetuning, by assigning return values to possible futures and sampling future embeddings based on their respective values.
Empirically, {\ours} outperforms or performs on par with its supervised pretraining counterpart, especially when dealing with sub-optimal data.
Further analysis reveals that {\ours} can extract diverse behaviors from offline data and controllably sample high-return behaviors by online finetuning.
Code is available at \href{https://github.com/fffffarmer/PDT}{here}.

\vspace{-4pt}

\end{abstract}

\section{Introduction}\label{sec:intro}

Large-scale pretraining has achieved phenomenal success in the fields of computer vision~\cite{chen2020simple,he2022masked} and natural language processing~\cite{devlin2018bert,radford2018improving}, where fast adaptation to various downstream tasks can be achieved with a unified model trained on a diverse corpus of data.
This trend has spurred interest in applying similar paradigms to reinforcement learning (RL), resulting in the prevalence of offline RL~\cite{levine2020offline,kumar2020conservative,agarwal2020optimistic}.
Offline RL aims to learn a task-solving policy exclusively from a static dataset of reward-labeled trajectories.
Given the resemblance between offline RL and supervised learning, recent research has further explored the direction of converting offline RL~\cite{chen2021decision,janner2021offline} or offline-to-online RL~\cite{zheng2022online} to a sequence modeling problem and using the expressive Transformer architecture~\cite{vaswani2017attention} for decision making.

At the heart of these Transformer-based approaches is the idea of conditioning policies on a desired outcome.
For example, Decision Transformer (DT, \citealt{chen2020simple}) learns a model to predict actions based on historical context and a target future return.
By associating decisions with future returns, DT can perform credit assignment across long time spans, showing strong performance on various offline tasks.
While promising, DT presents an incomplete picture as reward-labeled datasets are required to train return-conditioned policies.
In practice, task rewards are usually hard to access and poorly scalable to large-scale pretraining.
Besides, eschewing rewards during pretraining also allows the model to acquire generic behaviors that can be easily adapted for use in different downstream tasks.

In this work, we aim to equip DT with the ability to learn from reward-free and sub-optimal data.
Specifically, we consider the \textit{pretrain-then-finetune} scenario, in which the model is first trained on offline reward-free trajectories and then finetuned on the target task via online interactions.
To effectively pretrain a model, it must be able to extract reusable and versatile learning signals in the absence of rewards.
During finetuning, the model is required to quickly adapt to task rewards, which presents another challenge as to what learning signals can be aligned with rewards.

To address the above challenges, we propose an unsupervised RL pretraining method called Pretrained Decision Transformer ({\ours}).
Inspired by recent study of future-conditioned supervised learning~\cite{furuta2022generalized,venuto2022policy,yang2022dichotomy}, we enable {\ours} to condition on the more generalized future trajectory information for action prediction, which allows the model to learn from unsupervised offline data.
{\ours} jointly learns an embedding space of future trajectory as well as a future prior conditioned only on past information.
By conditioning action prediction on the target future embedding, {\ours} is endowed with the ability to reason over the future.
This ability is naturally task-agnostic and can be generalized to different task specifications.
To achieve efficient online finetuning in downstream tasks, one can easily retrofit the future-conditioned framework into return-conditioned supervised learning by associating each future embedding to its return.
This is realized by training a return prediction network to predict the expected return value for each future embedding, which can be justified from the views of controllable generation~\cite{Dathathri2020Plug} and successor features~\cite{barreto2017successor}.
At evaluation, {\ours} utilizes the learned future prior together with the return prediction network to controllably sample high-return futures.

We evaluate {\ours} on a set of Gym MuJoCo tasks from
the D4RL benchmark~\cite{fu2020d4rl}.
Compared with its supervised counterpart~\cite{zheng2022online}, {\ours} exhibits very competitive performance, especially when the offline data is far from expert behaviors.
Our analysis further verifies that {\ours} can: 1) make different decisions when conditioned on various target futures, 2) controllably sample futures according to their predicted returns, and 3) efficiently generalize to out-of-distribution tasks.

\begin{figure*}
    \centering
    \includegraphics[width=0.9\textwidth]{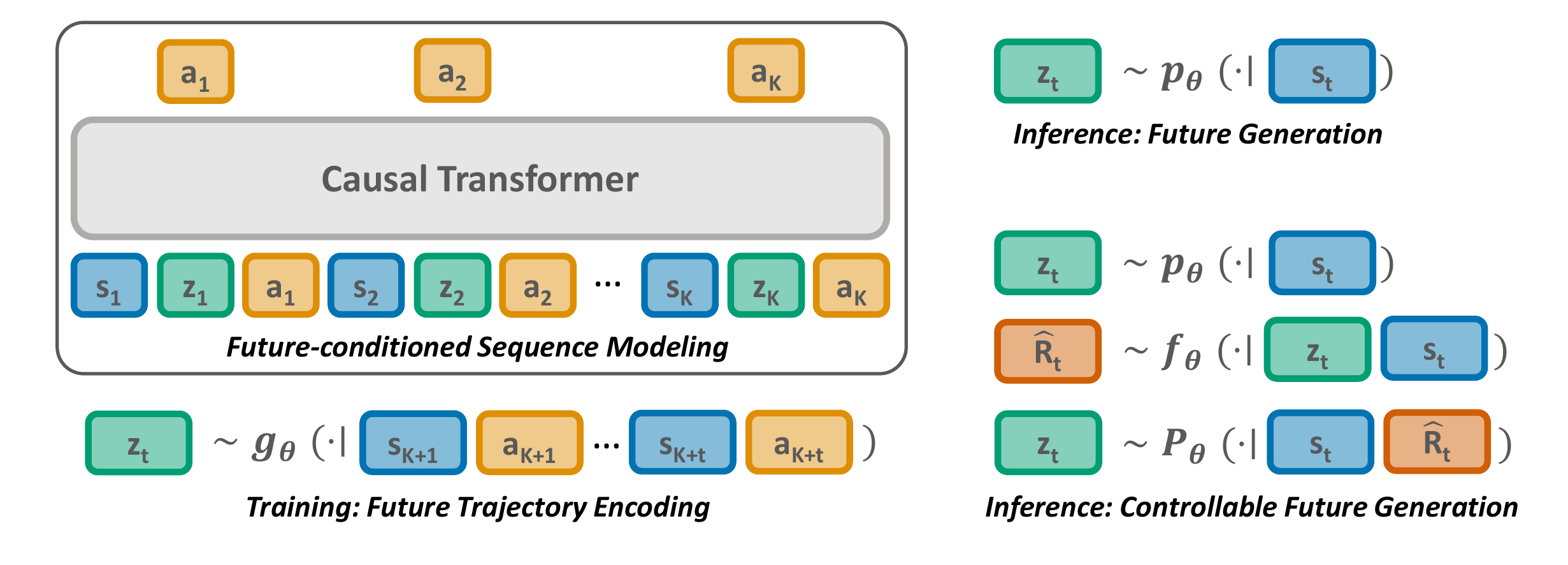}
    \vspace{-6pt}
    \caption{\textbf{Overview of the proposed {\ours} model.}
    {\ours} learns to make decisions based on future trajectory information.
    During training, future embeddings are extracted from the succeeding transitions by future encoder $g_\theta$.
    At evaluation, {\ours} samples future embeddings from future prior $p_\theta$ conditioned on the current state.
    When rewards are given in the finetuning phase, PDT learns a return predictor $f_\theta$ to steer the sampling procedure towards high-return future embeddings by Bayes' rule $P_\theta(z_t \mid \hat{R}_t, s_t) \propto p_\theta(z_t \mid s_t) f_\theta(\hat{R}_t \mid z_t, s_t)$.
    }
    \label{fig:illustration}
\end{figure*}
\section{Related Work}\label{sec:related}

Our work considers \textbf{future conditioning} as a powerful approach for \textbf{unsupervised RL pretraining}.
In this section, we review relevant works in these two research directions.

\subsection{Future-conditioned Supervised Learning}

Future-conditioned supervised learning has been gaining popularity recently in the field of RL due to its simplicity and competitiveness.
The idea is to predict actions conditioned on desired future outcomes, seeking to learn a policy with a future-conditioned supervised loss.
Among all kinds of outcomes, future rewards or returns ~\cite{schmidhuber2019reinforcement,kumar2019reward} are most commonly used.
\citet{chen2021decision} propose Decision Transformer (DT), a Transformer-based~\cite{vaswani2017attention} approach to learn a return-conditioned policy for offline RL.
\citet{zheng2022online} further extend DT to the offline-to-online RL setting by equipping DT with stochastic policies for online exploration.
\citet{emmons2022rvs} show that simple feed-forward MLPs are also capable to learn powerful future-conditioned policies.
However, Transformer-based approaches usually show good scaling properties~\cite{lee2022multigame}, in accord with results in the language domain~\cite{kaplan2020scaling}.

Other than target returns, future information such as goals~\cite{andrychowicz2017hindsight}, trajectory statistics~\cite{furuta2022generalized}, or learned trajectory embeddings~\cite{yang2022dichotomy,furuta2022generalized} can also be used to condition the policy.
Leveraging future information has been explored as means to combat environment stochasticity~\cite{villaflor2022addressing,yang2022dichotomy} or to improve value estimations for model-free RL~\cite{venuto2022policy}.
Although our approach potentially enjoys these advantages, in this work we are motivated in a different way and focus on how to tame future conditioning for unsupervised pretraining.


\subsection{Unsupervised Pretraining in RL}
Our work falls into the category of unsupervised pretraining in RL.
A number of works have sought to improve sample efficiency of RL by pretraining the agent with online interactions~\cite{eysenbach2018diversity,laskin2021urlb} or offline trajectories~\cite{ajay2021opal,schwarzer2021pretraining,juewu-mc} prior to finetuning on the target task.

Online unsupervised pretraining aims at learning generic skills by interacting with the environment. During pretraining, the agent is allowed to collect large-scale data from the environment without access to extrinsic rewards. To learn generic skills, existing methods use intrinsic rewards as a principled mechanism to encourage the agent to build its own knowledge. Based on how the intrinsic rewards are designed~\cite{xie2022pretraining}, prior works can be categorized into three classes: curiosity-driven exploration~\cite{burda2018exploration}, skill discovery~\cite{eysenbach2018diversity} and data coverage maximization~\cite{lee2019efficient}.

Although online unsupervised pretraining provides an effective framework to learn prior skills for downstream tasks, it requires abundant online samples which makes it sample inefficient. To address this issue, offline unsupervised pretraining has attracted attention. Existing works on offline unsupervised pretraining mainly fall into two classes. The first class is pretraining representation, aiming to learn good representation from large-scale offline data. To name a few, \citet{yang2021representation} propose attentive contrastive learning (ACL) which borrows the idea of BERT to pretraining representations that can be used to accelerate the behavior learning on downstream tasks. SGI~\cite{schwarzer2021pretraining} combines self-predictive representation, inverse dynamics prediction and goal-conditioned RL to learn powerful representations for visual observations. The second class is pretraining behavior. \citet{baker2022video} propose video pretraining (VPT), a semi-supervised scheme to utilize large-scale offline data without action information. Specifically, VPT learns an inverse dynamic model on a small-scale supervised data with action information and use the model to provide action for effective behavior cloning and finetuning. \citet{pertsch2021accelerating} propose a deep latent variable model that jointly learns an embedding space of skills and the skill prior from offline agent experience. 
Different from these works, we seek to address the issue of reward-centric optimization in existing approaches and enable efficient pretraining on large-scale unsupervised datasets.

\section{Preliminaries}\label{sec:preliminaries}

\subsection{Markov Decision Process}

Reinforcement learning (RL) typically models the environment as a Markov decision process (MDP), which can be described as $\left(\mathcal{S}, \mathcal{A}, P, r, \rho, \gamma\right)$, where $\mathcal{S}$ is the state space, $\mathcal{A}$ is the action space, $P\left(s_{t + 1} \mid s_t, a_t\right)$ is the probability distribution over transitions, $r$ is the reward function, $\rho$ is the initial state distribution, and $\gamma \in (0,1)$ is the discount factor.
The objective is to learn a policy that maximizes the cumulative return $\mathbb{E}\left[\sum_{t=1}^{T} r_t\right]$.

Let $\tau = \left(s_t, a_t\right)_{t=1}^T = \left(s_1, a_1, s_2, a_2, \cdots, s_T, a_T\right)$ denote a trajectory composed of a sequence of states and actions, and $\tau_{i: j}=\left(s_t, a_t\right)_{t=i}^j$ denote a sub-trajectory of $\tau$\footnote{$\tau_{t:t-1}$ represents an empty sub-trajectory.}.
Separating rewards from state-action dynamics is deliberate, as rewards are task-specific and usually human-provided.

\subsection{RL via Supervised Sequence Modeling}~\label{sec:rvs}

Previous work~\cite{chen2021decision,janner2021offline,lee2022multigame} has investigated casting offline RL as a sequential modeling task.
Specifically, Decision Transformer (DT, \citealt{chen2021decision}) takes the following trajectory representation as input, concatenating $\tau$ with rewards:
\begin{equation*}
    \hat{\tau}=(\hat{R}_1, s_1, a_1, \hat{R}_2, s_2, a_2, \ldots, \hat{R}_T, s_T, a_T),
\end{equation*}
where $\hat{R}_t = \sum_{t^{\prime}=t}^T r_{t^{\prime}}$, usually termed \textit{return-to-go} or simply \textit{return}, is the sum of future rewards from timestep $t$.
After tokenization, $\hat{\tau}$ is fed into a GPT-based Transformer~\cite{radford2018improving} which plays the role of an expressive policy function approximator $\pi_\theta$ to predict the next action\footnote{For simplicity, we use $\theta$ to represent all the learned parameters.}.
For offline RL, the policy is trained to maximize the likelihood of actions in the reward-labeled offline dataset $\hat{\mathcal{D}} = \left\{\hat{\tau}^{(m)}\right\}_{m=1}^M$:

\begin{equation*}~\label{eq:dt}
    \mathcal{L}_{\mathrm{DT}}=\mathbb{E}_{\hat{\tau} \sim \hat{\mathcal{D}}}\left[\sum_{t=1}^T-\log \pi_\theta(a_t \mid \hat{\tau}_{1: t-1}, s_t, \hat{R}_t)\right].
\end{equation*}

To enable online finetuning, \citet{zheng2022online} further propose Online Decision Transformer (ODT), equipping DT with a stochastic policy and training the model via an additional max-entropy objective:

\begin{equation*}
    \mathcal{L}_{\mathrm{ODT}} = \mathcal{L}_{\mathrm{DT}} -\alpha \mathbb{E}_{\hat{\tau} \sim \hat{\mathcal{D}}}\left[\sum_{t=1}^{T} H(\pi_\theta(\cdot \mid \hat{\tau}_{1: t-1}, s_t, \hat{R}_t))\right],
\end{equation*}
where $\alpha$ is the temperature parameter~\cite{haarnoja2018soft} to control its stochasticity.

While return-conditioned approaches like DT and ODT have gained great attention due to their simplicity and effectiveness when applying to offline RL problems, it requires reward signals to train, which is naturally hard to scale to large-scale pretraining.
Besides, using a single scalar value (i.e., the target return) as input could fail to capture sufficient future information.
These deficiencies motivate us to design new pretraining algorithms.








\subsection{Unsupervised Pretraining}
We consider the unsupervised pretraining regime where the goal is to leverage easy-to-collect offline data free of rewards for more data-efficient reinforcement learning.

In the pretraining phase, we assume reward-free offline dataset $\mathcal{D} = \left\{\tau^{(m)}\right\}_{m=1}^M$is available.
Similar to what we have witnessed in the fields of computer vision~\cite{chen2020simple,he2022masked} and natural language processing~\cite{devlin2018bert,radford2018improving}, it is a promising direction to leverage highly sub-optimal and unlabeled training data to conduct RL pretraining. 

In the downstream phase, rewards associated with a task are revealed to the agent.
We expect the agent to quickly adapt to the task by reusing its prior knowledge learned in the pretraining phase.
While offline finetuning is also reasonable, 
in this work we focus on online finetuning as it requires agents to trade off exploration and exploitation and is usually considered harder~\cite{zheng2022online}.

\section{Methodology}~\label{sec:method}


In this section, we first describe the future conditioning framework that scaffolds {\ours}.
We then present how to pretrain {\ours} in an unsupervised manner and use the pretrained model for efficient task adaptation. Finally, we show the connection between the ideas of {\ours} and successor features~\cite{barreto2017successor}.
Figure~\ref{fig:illustration} gives an overview of the proposed {\ours}.






\subsection{Learning to Act by Incorporating the Future}

As discussed earlier in Section~\ref{sec:rvs}, return-conditioned approaches are greatly restricted by rewards.
To eliminate the need for reward information, we draw inspiration from future-conditioned RL~\cite{furuta2022generalized,emmons2022rvs} and use it to ground unsupervised pretraining.

Specifically, we condition the policy on future latent variables instead of target returns.
Let $g_{\theta}$ be a trajectory-level future encoder and $\pi_{\theta}$ be a policy network.
We condition $\pi_\theta$ on future latent variables $z \sim g_{\theta}(\cdot \mid \tau)$, where $g_{\theta}(\cdot \mid \tau)$ outputs a multivariate Gaussian distribution.
We expect the model to capture different behaviors seen in the offline data during unsupervised pretraining.
These task-agnostic behaviors can be combined to form different policies based on task-specific information (i.e., rewards).
Therefore, when the pretrained model is applied to downstream tasks, the agent is able to control which behaviors to sample based on the provided reward information. 
At test time, the algorithm takes the learned policy $\pi_{\theta}$ along with a learned prior $p_{\theta}\left(z \mid s_t\right)$ to take actions:

\begin{equation*}
    a_t \sim \pi_{\theta}\left(\cdot \mid \tau_{1: t-1}, s_t, z\right), z \sim p_{\theta}\left(\cdot \mid s_t\right).
\end{equation*}

The framework of future conditioning brings several benefits.
Firstly, it allows us to disentangle rewards from target outcomes, opening up opportunities for large-scale unsupervised pretraining.
Secondly, it alleviates the issue of inconsistent behaviors induced by return-conditioned supervised learning, where behaviors significantly deviate from the intended targets~\cite{yang2022dichotomy}.

\subsection{Future-conditioned Pretraining}

In this work, similar to that used in DT~\cite{chen2021decision,zheng2022online}, we use a GPT-based Transformer architecture~\cite{radford2018improving} to parameterize the policy network.
Given an input sequence $(s_1, a_1, s_2, a_2, \ldots, s_T, a_T)$, the behavior cloning objective resembles that of ODT:

\begin{equation}~\label{eq:main_dt}
    \begin{aligned}
        \mathcal{L}_\mathrm{BC} &= \mathbb{E}_{\substack{\tau \sim \mathcal{D}\\z \sim g_\theta(\cdot \mid \tau)}}
        \left[\sum_{t=1}^{T}-\log \pi_\theta(a_t \mid \tau_{1: t-1}, s_t, z)\right] \\
        &-\alpha \mathbb{E}_{\substack{\tau \sim \mathcal{D}\\z \sim g_\theta(\cdot \mid \tau)}}\left[\sum_{t=1}^{T} H(\pi_\theta(\cdot \mid \tau_{1: t-1}, s_t, z)\right],
    \end{aligned}
\end{equation}
where $H(\pi_{\theta})$ denotes the entropy of action distribution, and $\alpha$ represents a hyperparameter that trades off the contribution of entropy maximization. Note that here we apply the reparameterization trick to allow gradients to backpropagate through the future encoder.


Training the future encoder with the above objective allows us to efficiently leverage future trajectory information to predict actions. 
However, without explicit regularization, the future encoder can collapse and fail to capture the full distribution of future information. Besides, during execution, we need a prior to guide the agent to sample from the future embedding space. 
Therefore, inspired by previous work~\cite{ajay2021opal,pertsch2021accelerating}, we use the following objective to train the future encoder and a prior:

\begin{equation}~\label{eq:future}
    \begin{aligned}
        \mathcal{L}_\mathrm{future}&=\beta\mathbb{E}_{\tau \sim \mathcal{D}}\left[D_{\mathrm{KL}}\left(g_{\theta}(z \mid \tau) \| \mathcal{N}(\mathbf{0}, I)\right)\right] \\
        &+\mathbb{E}_{\tau \sim \mathcal{D}}\left[D_{\mathrm{KL}}\left(\lfloor g_{\theta}(z \mid \tau)\rfloor \| p_{\theta}(z \mid s_t)\right)\right],
    \end{aligned}
\end{equation}
where $D_{\mathrm{KL}}$ denotes the Kullback–Leibler (KL) divergence, $\lfloor \cdot \rfloor$ denotes the stop-gradient operator and $\beta$ is a hyperparameter.
The former serves as a regularization term on the future encoder to constrain the capacity of the latent $z$~\cite{higgins2017betavae}.
The second term is applied to learn a prior model $p_\theta(z|s_t)$ which helps sample behaviors based on the distribution of offline data.
In Appendix~\ref{appendix:opal}, we discuss the differences between {\ours} and OPAL~\cite{ajay2021opal} in terms of their learning objectives.

\begin{algorithm}[t]
\caption{Future-conditioned Pretraining}
\label{algo:pretrain}
\begin{algorithmic}[1]
\INPUT policy $\pi_\theta$, future prior $p_\theta$, future encoder $g_\theta$, reward-free offline dataset $\mathcal{D}$, context length $K$, batch size $B$, training iteration $I$

\FOR{$i = 1, \ldots, I$}
    \STATE Sample $B$ trajectories from $\mathcal{D}$ according to $p(\tau)=|\tau| / \sum_{\tau^\prime \in \mathcal{D}}|\tau^\prime|$

    \FOR{each sampled trajectory $\tau$}
        \STATE $\tau_{t:t+K-1} \gets$ a length-$K$ sub-trajectory uniformly sampled from $\tau$
        \STATE $\tau_{t+K:t+2K-1} \gets$ a length-$K$ future sub-trajectory
        \STATE Sample $z \sim g_{\theta}(\cdot | \tau_{t+K:t+2K-1})$
        \STATE Predict actions $a_{t^\prime} \sim \pi_\theta(\cdot \mid \tau_{t: t^\prime-1}, s_{t^\prime}, z)$, $\forall t^\prime = t, \ldots, t + K - 1$
        \STATE Calculate $\mathcal{L} = \mathcal{L}_\mathrm{BC} + \mathcal{L}_\mathrm{future}$ (Equation~\ref{eq:main_dt},\ref{eq:future})
    \ENDFOR
    \STATE Update $\theta$ by gradient descent
\ENDFOR

\end{algorithmic}
\end{algorithm}

Algorithm~\ref{algo:pretrain} summarizes the future-conditioned pretraining procedure for {\ours}.
In practice, {\ours} takes a sub-trajectory as the history context and encodes the succeeding sub-trajectory to obtain the future embedding.
Intuitively, we empower the model to take actions based on future trajectory information.
This allows {\ours} to learn robust and general behaviors in an unsupervised manner.




\subsection{Return-conditioned Finetuning}~\label{sec:finetuning}

Albeit useful to sample future latent variables and generate behaviors imitating the distribution of offline data, $p_\theta(z \mid s_t)$ fails to encode any task-specific information.
Therefore, it is required to steer $p_\theta(z \mid s_t)$ to sample futures embeddings that lead to high future return during finetuning.

This leads to controllable generation, an active area of research in computer vision~\cite{nie2021controllable,dhariwal2021diffusion} and natural language processing~\cite{Dathathri2020Plug}.
\citet{lee2022multigame} consider applying controllable generation to generate expert behaviors for return-conditioned DT.
In contrast to controlling a return-conditioned policy by assigning a scalar target return, we need to address a more challenging problem of assigning credits to the future.
By Bayes' rule, we have:
\begin{equation*}
    p(z \mid \hat{R}_t, s_t) \propto p(z \mid s_t) p(\hat{R}_t \mid z, s_t),
\end{equation*}
which suggests that we can sample the desired high-return future embedding by steering the future prior with $p(\hat{R}_t \mid z, s_t)$.
Since this distribution is unknown, we use a return prediction network $f_\theta(\cdot \mid z, s)$ to predict $p(\hat{R}_t \mid z, s_t)$.
We parameterize $p(\hat{R}_t \mid z, s_t)$ as a Gaussian distribution with learned mean and variance.
The return prediction network is trained along with all the other objectives during finetuning:

\begin{equation} \label{eq:return}
    \mathcal{L}_\mathrm{return} = \mathbb{E}_{\substack{\hat{\tau} \sim \hat{\mathcal{D}} \\ z \sim g_{\theta}(\cdot \mid \hat{\tau})}}\left[-\log f_\theta\left(\hat{R}_t \mid z, s_t\right)\right].
\end{equation}

Similar to Equation~\ref{eq:main_dt}, we apply the reparameterization trick for the future encoder $g_{\theta}$ in Equation~\ref{eq:return}.
This allows gradients to backpropagate to the future encoder, regularizing it to encode task-specific reward information during finetuning.

In practice, we warm-up the return predictor with online exploration transitions at the beginning of the finetuning phase.
We consider directly sampling high-return futures, rather than conditioning on a high target return.
Specifically, a batch of future embeddings is randomly sampled from the future prior model $p_\theta$ and the one with the highest predicted return is selected to condition $\pi_\theta$ during inference, which eliminates the need for a pre-determined target return.
Algorithm~\ref{algo:finetune} summaries the finetuning procedure for {\ours}.

\begin{table*}[t]
    \centering

\scriptsize
\begin{tabular}{l|lll|lllll|lll}
\thickhline
dataset                   & Mean   & Min   & Max    & SAC                                     & ACL            & {\ours}-0      & {\ours}                & $\delta_\text{\ours}$            & ODT-0                            & ODT            & $\delta_\text{ODT}$             \\ \hline
hopper-m             & 44.32  & 10.33 & 99.63  & 24.22 {\tiny $\pm$ 10.55}                          & 57.66 {\tiny $\pm$ 6.23} & 53.74                           & \textbf{95.26} {\tiny $\pm$ 1.77} & \textbf{41.52}  & 66.01                            & 87.22 {\tiny $\pm$ 6.85}  & 21.22                           \\
hopper-m-r      & 14.98  & 0.58  & 98.73  &                                         & 51.68 {\tiny $\pm$ 48.74} & 28.56                           & \textbf{84.96} {\tiny $\pm$ 5.49} & \textbf{56.40}  & 74.36                            & 75.31 {\tiny $\pm$ 6.22}  & 0.95                            \\
walker2d-m           & 62.09  & -0.18 & 92.04  & 35.26 {\tiny $\pm$ 23.51}                          & 60.21 {\tiny $\pm$ 27.08} & 73.70 & \textbf{75.24} {\tiny $\pm$ 4.60}                           & \textbf{1.53}                            & 72.80                            & 72.62 {\tiny $\pm$ 5.51}  & -0.18 \\
walker2d-m-r    & 14.84  & -1.13 & 89.97  &                                         & \textbf{87.54} {\tiny $\pm$ 7.31}  & 15.64                           & 58.58 {\tiny $\pm$ 14.78}                          & \textbf{42.94}  & 73.27                            & 70.54 {\tiny $\pm$ 2.89}  & -2.73                           \\
halfcheetah-m        & 40.68  & -0.24 & 45.02  & \textbf{57.05} {\tiny $\pm$ 3.89} & 46.59 {\tiny $\pm$ 2.71}  & 42.86                           & 37.93 {\tiny $\pm$ 1.82}                           & \textbf{-4.93}  & 42.69                            & 35.07 {\tiny $\pm$ 10.40} & -7.62                           \\
halfcheetah-m-r & 27.17  & -2.89 & 42.41  &                                         & 50.56 {\tiny $\pm$ 3.74}  & 24.83                           & 29.70 {\tiny $\pm$ 4.97}                           & \textbf{4.88}   & 40.95                            & 35.60 {\tiny $\pm$ 1.68}  & -5.35                           \\
ant-m                & 80.30  & -4.85 & 107.31 & 33.30 {\tiny $\pm$ 12.10}                          & 28.44 {\tiny $\pm$ 10.78}  & \textbf{93.86} & 89.10 {\tiny $\pm$ 6.49}                           & \textbf{-4.77}  & 93.08                            & 73.80 {\tiny $\pm$ 16.77} & -19.28                          \\
ant-m-r         & 30.95  & -8.87 & 96.56  &                                         & 9.53 {\tiny $\pm$ 1.80}   & 53.78 & 48.18 {\tiny $\pm$ 9.59}                           & \textbf{-5.60}  & \textbf{90.37}                            & 60.48 {\tiny $\pm$ 6.23}  & -29.89                          \\ \hline
sum                       & 315.33 & -7.25 & 671.67 &                                         & 392.22         & 386.96                          & 518.93                                  & \textbf{123.94} & \textbf{553.53} & 510.65         & -42.87\\
\thickhline
\end{tabular}

    \caption{
    \textbf{Performance on Gym MuJoCo tasks.}
    We run each instance for 200k online transitions, and measure the finetuning performance by the averaged normalized return over 3 random seeds.
    We also report the zero-shot performance of the pretrained model with suffix ``-0''.
    $\delta_\textrm{x}$ shows the performance gain during online finetuning.
    The best performance for each task is highlighted in \textbf{bold}.
    }
    
    \label{tab:mujoco}
\end{table*}

\subsection{Connection to Successor Features}

As discussed in the previous section, our approach assigns return values to future embeddings for fast task adaptation.
This bears resemblance to successor features (SFs, \citealt{barreto2017successor}), a framework for transfer learning in RL.
SFs assume that the one-step rewards can be written as:

\begin{equation}~\label{eq:sf}
    r\left(s, a\right)=\phi\left(s, a\right)^{\top} \mathbf{w},
\end{equation}
where $\phi\left(s, a\right)$ represents the task-agnostic dynamics of the environment whereas $\mathbf{w}$ specifies the task preference.
Based on this assumption, the action-value function widely applied in model-free RL algorithms is given by $Q^{\pi}(s, a)=\psi^\pi(s, a)^{\top} \mathbf{w}$, where $\psi^\pi(s, a) = \mathbb{E}^\pi\left[\sum_{t=0}^{\infty} \gamma^t \phi\left(s_{t+1}, a_{t+1}\right) \mid s_t=s, a_t=a\right]$ is the SF summarizing the dynamics induced by $\pi$ in the future.

We establish the connection between {\ours} and SFs by applying the assumption in Equation~\ref{eq:sf} to returns: 

\begin{equation*}
    \hat{R}_{t}=\sum_{t^{\prime}=t}^T r_{t^{\prime}}=\left[\sum_{t^{\prime}=t}^T\phi(s_{t^
    \prime+1}, a_{t^
    \prime+1})\right]^\top \mathbf{w}.
\end{equation*}
Here, feature encoder $\phi$ needs to be determined, and our future encoder $g_\theta$ can be a good candidate to directly encode the summation.
Since $g_\theta$ is pretrained on reward-free offline trajectories, it  naturally encodes information about the environment dynamics.
This allows fast task adaptation in the downstream finetuning phase, which is achieved by learning task-specific $\mathbf{w}$ with the return prediction network.

\section{Experiments}~\label{sec:exp}

We conduct several experiments to ascertain the effectiveness of {\ours}, with the aim to gain insights into the following:
\begin{itemize}
    \item Can {\ours} extract rich prior knowledge from reward-free offline data to facilitate downstream learning?
    \item Is unsupervised pretraining comparable with supervised pretraining?
    Does {\ours} achieve better generalization than its supervised counterpart?
    \item How do future conditioning and controllable sampling respectively contribute to {\ours}'s performance?
\end{itemize}

\subsection{Baselines}
We compare the performance of {\ours} to several competitive baselines, including both unsupervised and supervised pretraining methods.
\begin{itemize}

    \item \textbf{Soft Actor-Critic} (\textbf{SAC}, \citealt{haarnoja2018soft}) trains a off-policy agent from scratch.
    We include this baseline to test the benefit of leveraging prior experience.



    \item \textbf{Attentive Contrastive Learning} (\textbf{ACL}, \citealt{yang2021representation}) considers various representation learning objectives to pretrain state representations on offline data.
    The pretrained representations are then combined with SAC to solve downstream tasks.

    \item \textbf{Online Decision Transformer} (\textbf{ODT}, \citealt{zheng2022online}) pretrains a stochastic return-conditioned policy using reward-labeled offline data.
    This baseline serves as a supervised counterpart of {\ours}, and we include it to test whether supervised pretraining is superior to the proposed unsuperivsed method.
\end{itemize}

For SAC, we use the open-source codebase\footnote{\url{{https://github.com/denisyarats/pytorch_sac}}} and the hyperparameters in \citet{haarnoja2018soft}.
For ACL, we use the official implementation\footnote{\url{https://github.com/google-research/google-research/tree/master/rl_repr}}, and choose the most competitive reward-free variant and the hyperparameters in \citet{yang2021representation}.
We use the official implementation\footnote{\url{https://github.com/facebookresearch/online-dt}} for ODT.
Our {\ours} implementation is based on the ODT codebase.
Please see Appendix~\ref{appendix:hyperparameter} for more details.

\subsection{Benchmark Datasets}
We evaluate our method on the Gym MuJoCo datasets from D4RL~\cite{fu2020d4rl}.
These datasets consiste of offline trajectories collected by partially trained policies in four simulated locomotion domains: halfcheetah, hopper, walker2d,
and ant.
Since we care most about how well {\ours} learns from sub-optimal data, we choose the medium and medium-replay datasets whose trajectories are far from task-solving.

Different from offline RL, unsupervised pretraining assumes that rewards are not available to agents when pretraining on the offline data.
After the pretraining phase, agents are allowed to interact with the environment, finetuning their policy from its own behaviors and the corresponding rewards.
Following \citet{zheng2022online}, we consider a relatively small budget of 200k online interactions.
This requires the agent to learn from unlabeled data effectively and adapt to downstream tasks quickly.
While \citet{zheng2022online} use offline pretraining data to initialize the replay buffer for finetuning, it is infeasible for the considered baselines to utilize reward-free data.
Therefore, we adopt the standard protocol where the agent uses its own rollouts to initialize the replay buffer for finetuning.


\subsection{Gym MuJoCo}

\begin{figure*}[t]
    \centering

    \begin{subfigure}[b]{0.245\linewidth}
         \centering
         \includegraphics[width=\linewidth]{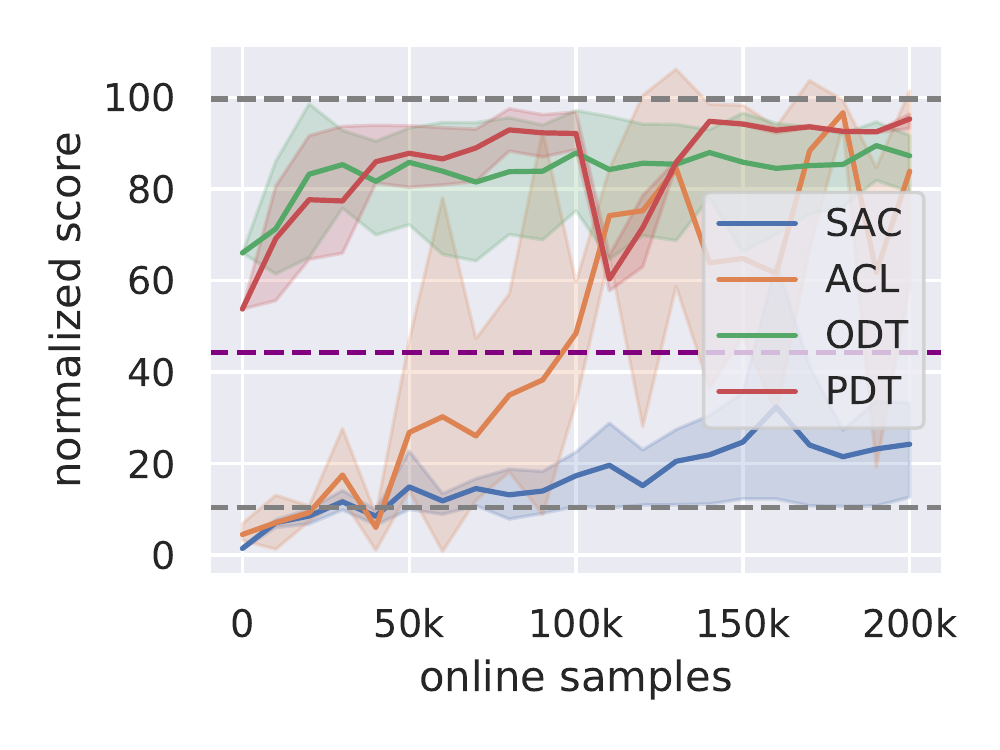}
         \caption{hopper-medium}
    \end{subfigure}
    \begin{subfigure}[b]{0.245\linewidth}
         \centering
         \includegraphics[width=\linewidth]{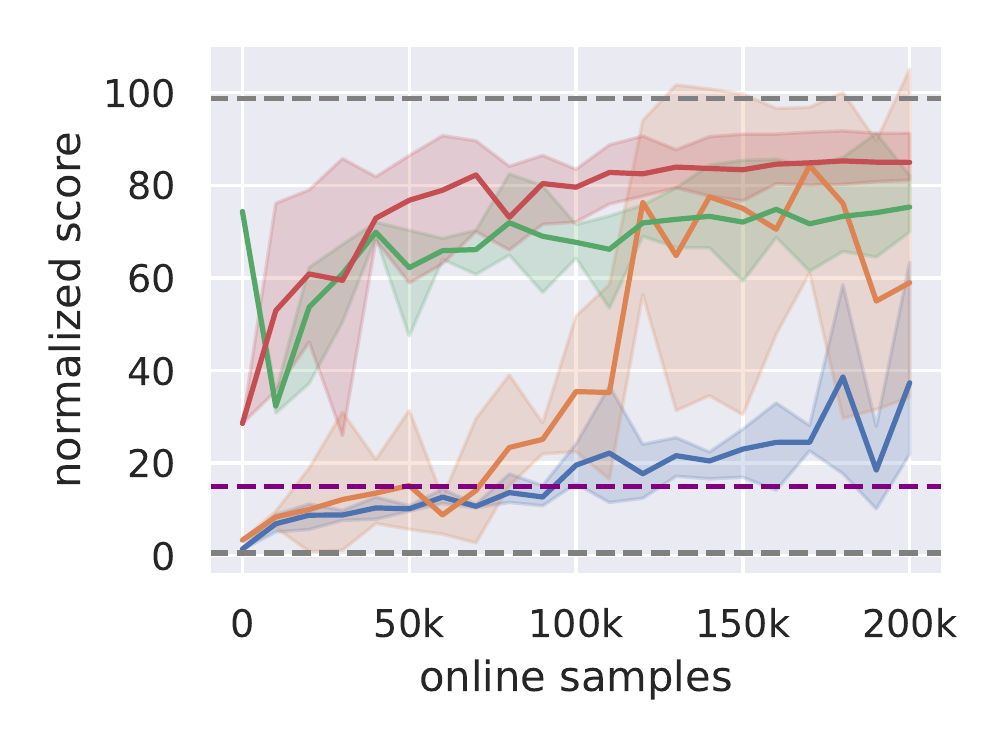}
         \caption{hopper-medium-replay}
     \end{subfigure}
    \begin{subfigure}[b]{0.245\linewidth}
         \centering
         \includegraphics[width=\linewidth]{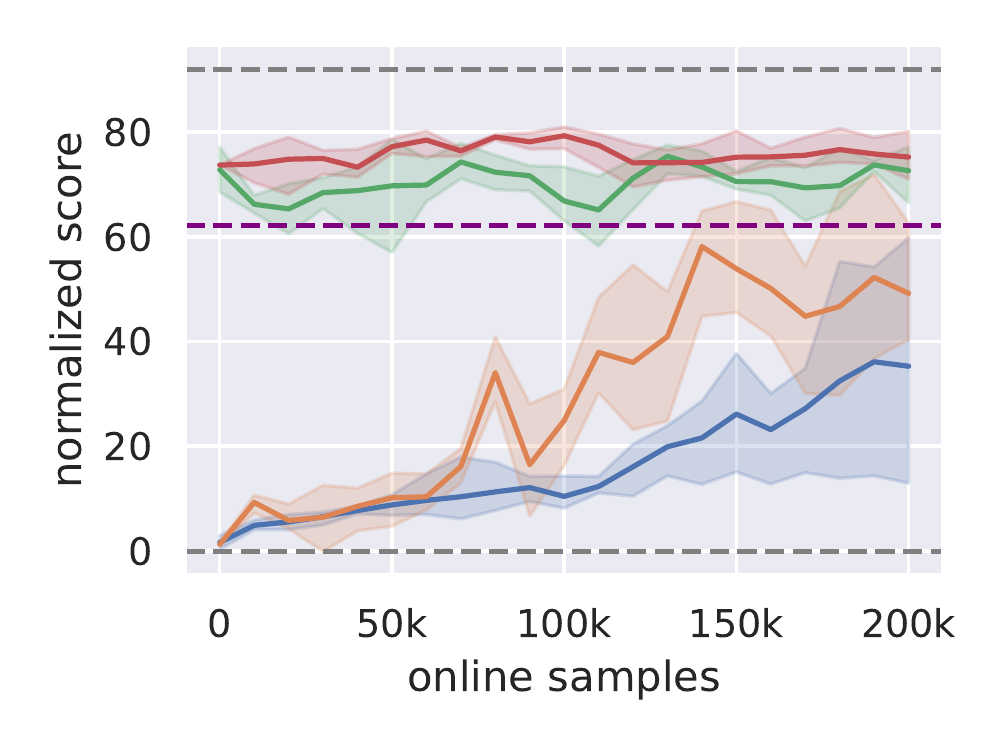}
         \caption{walker2d-medium}
    \end{subfigure}
    \begin{subfigure}[b]{0.245\linewidth}
         \centering
         \includegraphics[width=\linewidth]{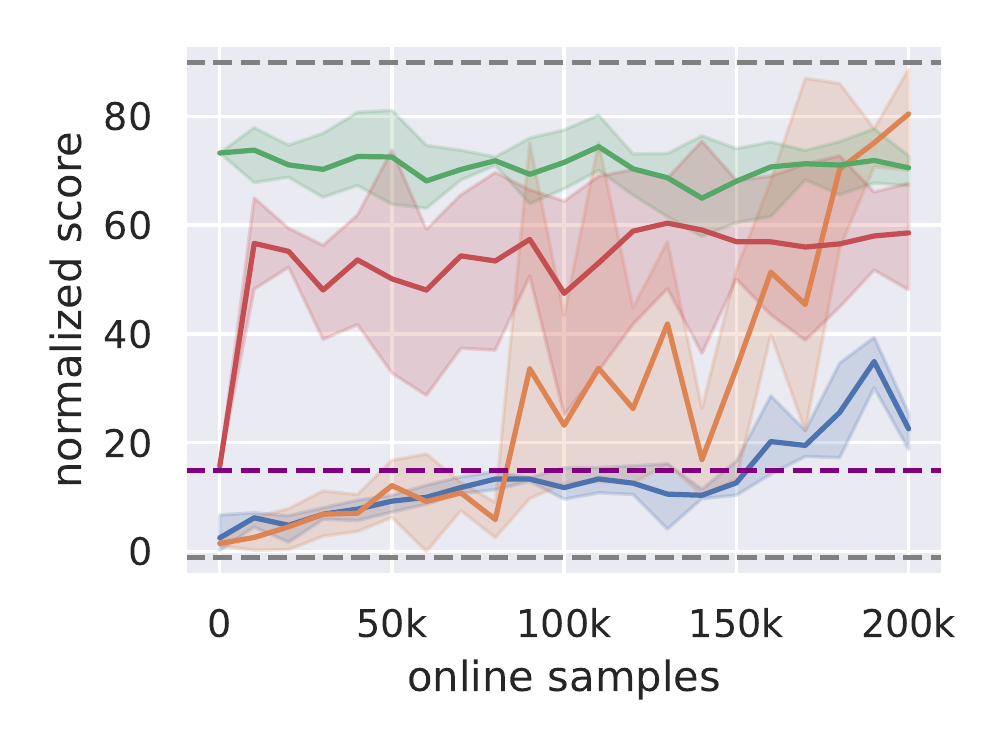}
         \caption{walker2d-medium-replay}
     \end{subfigure}
    \caption{
    \textbf{Downstream performance on Gym MuJoCO tasks.}
    Each instance is evaluated over 10 episodes every 10k transitions.
    Results are averaged over 3 random seeds.
    The shaded area shows the 95\% confidence interval.
    The \textcolor{mypurple}{purple} dashed line represents the averaged episodic return of offline trajectories, whereas the dashed lines in \textcolor{mygray}{gray} represent the min and max values.
    }
    \label{fig:mujoco_return}
\end{figure*}

We train all instances with 3 random seeds.
At evaluation, we run the policy for 10 episodes and compute the average return.
The reported results are normalized according to \citet{fu2020d4rl}.
Table~\ref{tab:mujoco} summaries the main results and Figure~\ref{fig:mujoco_return} shows the evaluation results during finetuning.

In comparison to learning from scratch using SAC, all the pretraining approaches improve sample efficiency substantially.
A closer look at the finetuning phase suggests that {\ours} not only shows good initial performance by learning to act based on the future, but also quickly adapts to the target task under the supervision of task rewards.
Due to the lack of prior knowledge, SAC starts with poor initial performance and heavily relies on online exploration to collect transitions, which results in relatively low sample efficiency.
The results of ACL suggest that pretrained representations can provide improvements in terms of sample efficiency in most cases, but we also observe the phenomenon of negative transfer on ant-medium-replay where ACL performs far behind SAC.
Besides, since ACL only pretrains state representations, it lacks the ability to reuse behaviors that can potentially benefit the target task.

Perhaps a more interesting observation is that {\ours} performs on par with its supervised counterpart ODT.
If we only consider offline pretraining, ODT outperforms {\ours} substantially with the help of supervision.
But when finetuning, {\ours} exhibits better sample efficiency, especially when the model is pretrained on the replay datasets  (e.g., hopper-medium-replay) composed of highly sub-optimal trajectories.
This suggests that {\ours} can quickly associate futures with return values and latch on high-return behaviors for data-efficient finetuning.

To support that {\ours} achieves non-trivial performance, we also compare {\ours} with Rewardless-DT, a simple baseline that builds on ODT but masks out return embeddings during pretraining.
Figure~\ref{fig:rewardless-dt} shows the finetuning performances of the sequential modeling methods.
We observe that Rewardless-DT exhibits relatively good performance on the medium datasets.
However, it struggles on the medium-replay datasets. 
This demonstrates that {\ours} can extract reusable learning signals from unsupervised pretraining, particularly in cases where the offline data is sub-optimal.




\subsection{Analysis}

In this section, we seek to gain insight into the key components of {\ours}, including future conditioning and controllable sampling.
Out investigation focuses on how the choice of future embeddings impacts generation, as well as the effect of regularization on future embeddings.
We also conduct an evaluation to test whether {\ours} exhibits better generalization capabilities as compared to its supervised counterpart.

\paragraph{Future conditioning.}~\label{sec:exp-future}
Ideally, future conditioning enables {\ours} to behave differently based on different target futures.
The ability to reason about the future can be very beneficial, especially at the early stage of an episode when the agent has few history transitions to ground decisions.

\begin{figure}[t]
    \centering
    \includegraphics[width=\linewidth]{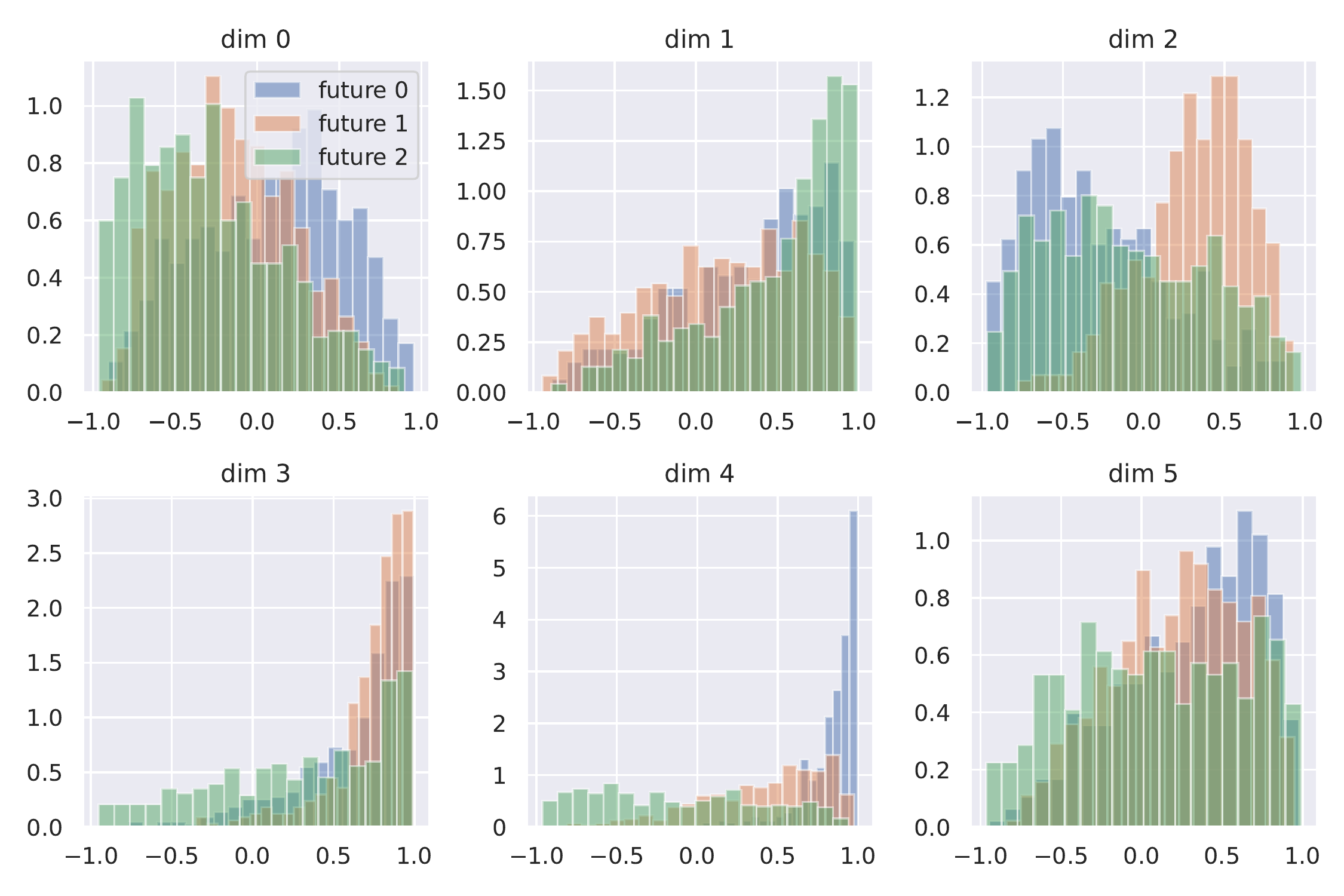}
    \caption{
    \textbf{Future conditioning} enables diverse behavior generation.
    We plot the histogram of action distribution produced by {\ours} for each dimension
    of the action space.
    The result is taken from the initial state of an episode when running {\ours} pretrained on the walker2d-medium dataset.
    See Appendix~\ref{appendix:future} for details.
    }
    \label{fig:pretraining}
\end{figure}

Figure~\ref{fig:pretraining} shows the action distributions given by a pretrained {\ours} agent at the initial state of an episode, conditioned on three different future embeddings sampled from the future prior.
We can observe that {\ours} makes clearly different decisions in response to the choice of target future.
Figure~\ref{fig:appendix-pretrain} further demonstrates that, as the episode continues, {\ours} relies more on its history transitions to make decisions whereas future information plays a less important role.

\paragraph{Performance vs. regularization.}

\begin{figure}[b]
    \centering
    \includegraphics[width=\linewidth]{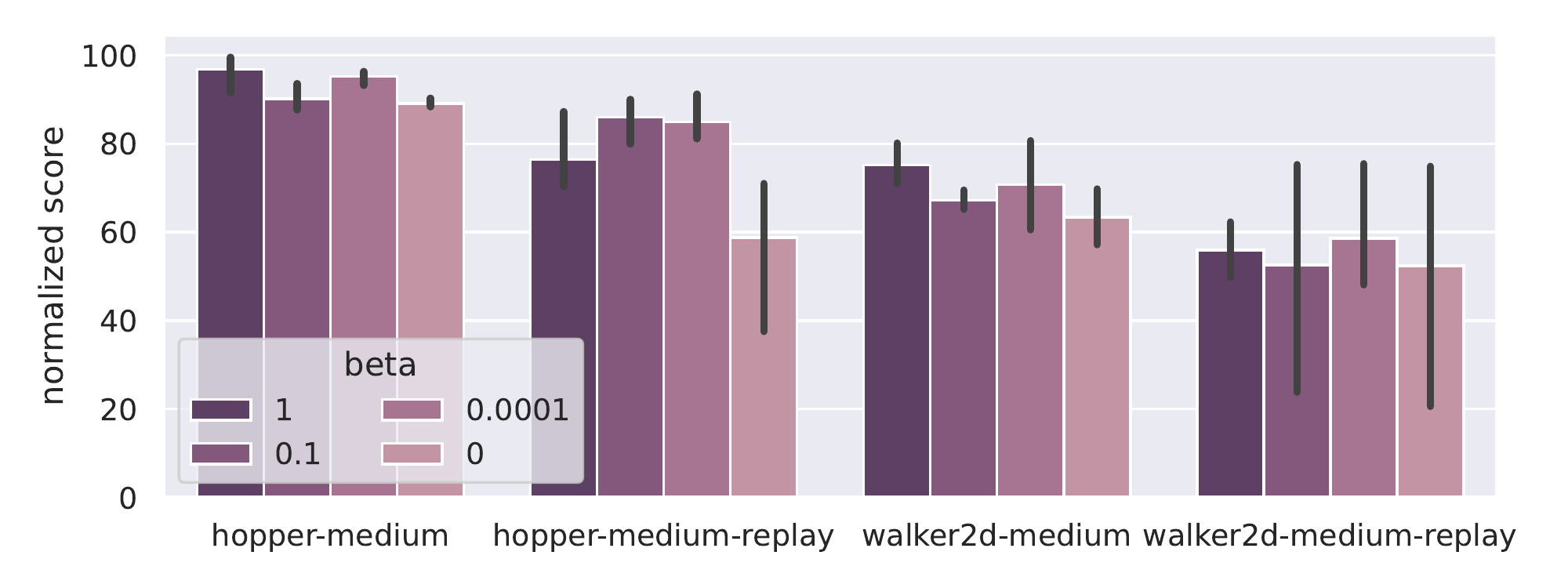}
    \caption{
    \textbf{Ablation results of future regularization} for downstream performance.
We run each instance for 200k online transitions and report the averaged normalized return over 3 random seeds. The error bar shows the 95\% confidence interval.
}
    \label{fig:finetunekl}
\end{figure}

$\beta$, which controls the regularization strength over future embeddings, is an important hyperparameter for {\ours}.
Intuitively, a large $\beta$ value discourages the policy to learn diverse behaviors, as minimizing the KL divergence limits how much future information is contained in the latent vectors.
In contrast, if $\beta$ is too small, {\ours} tends to overly rely on the privileged future information, in the sense that the latent vectors will contain sufficient information so that {\ours} ignores its own past.
Figure~\ref{fig:finetunekl} shows that, when pretrained on the medium datasets, {\ours} usually achieves better finetuning performance with larger $\beta$.
When pretrained on the medium-replay datasets, {\ours} favors a smaller $\beta$ value for finetuning.

\begin{figure}
    \centering
    \includegraphics[width=\linewidth]{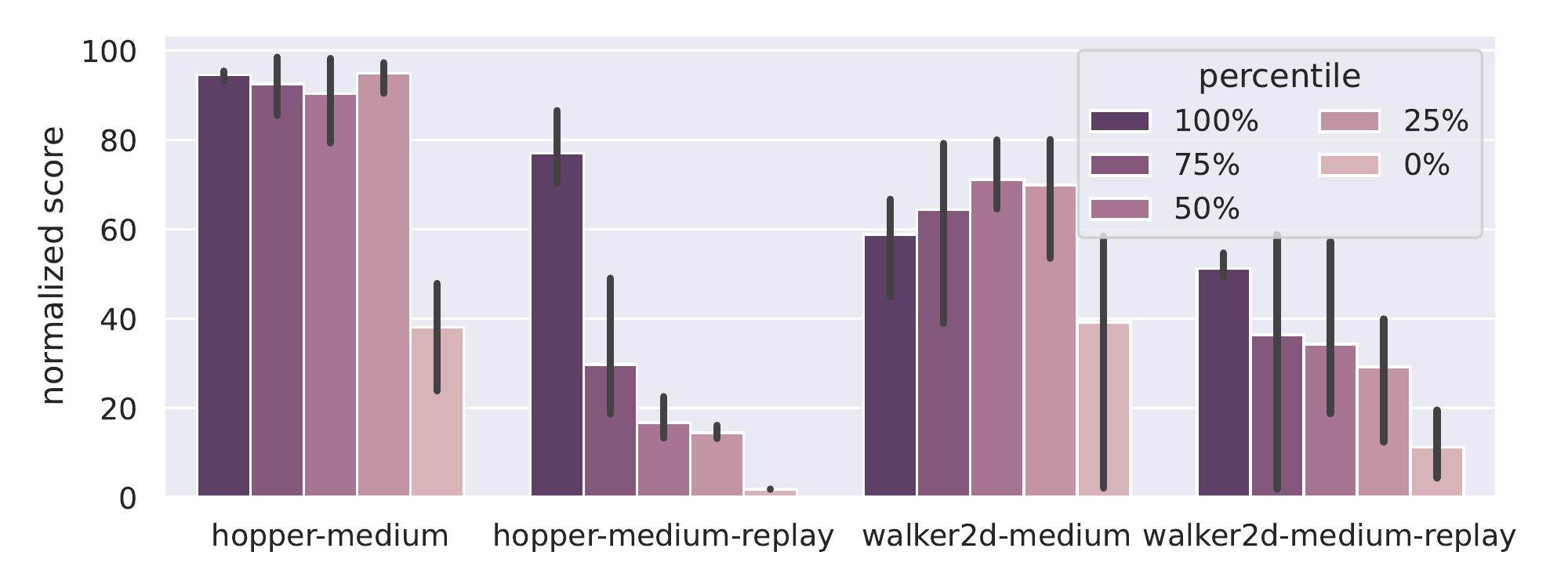}
    \caption{
    \textbf{Ablation of controllable sampling} for downstream performance.
    We condition {\ours} on future embeddings with different percentiles of predicted return.
    We run each instance for 250 online exploration episodes and report the averaged normalized return over 3 random seeds. The error bar shows the 95\% confidence interval.
    See Appendix~\ref{appendix:sampling} for more details.
    }
    \label{fig:sampling}
\end{figure}

\paragraph{Behavior diversity vs. regularization.}
To further investigate the connection between the strength of regularization and the diversity of generated behaviors, we conduct another set of experiments.
For each {\ours} policy, we evaluate its behavior diversity by how action distributions vary with different future embeddings to condition the policy.
The more dissimilar the action distributions are, the more diverse behaviors {\ours} can generate by sampling different future embeddings.

As shown in Figure~\ref{fig:diversity}, $\beta$ = 1 consistently leads to less diverse behaviors.
Besides, we observe that phenomenon becomes more pronounced on {x}-medium-replay datasets, indicating that we can regulate the behavior diversity exhibited by {\ours} when dealing with diverse data.

\paragraph{Controllable sampling.}
To steer the pretrained model to perform certain behaviors as specified by the downstream task, {\ours} uses a learned prior of future together with a return prediction network to sample high-return target futures as contexts to take actions.
Hence, it is crucial for the return prediction network to faithfully associate futures with their ground-truth returns.
To evaluate this, we condition {\ours} on the $x$-percentile of the predicted returns during online finetuning (e.g., $x=100\%$ recovers the original strategy), and varies the choice of $x$.

Figure~\ref{fig:sampling} shows the performance of {\ours} after 250 online exploration episodes.
We can observe that {\ours} achieves varied outcomes when conditioned on different target futures.
This again verifies our findings in our former analysis that {\ours} relies heavily on target future information to make decisions.
Besides, the results demonstrate that the return prediction network is effective in filtering high-return futures out of all the possibilities.
This property is pivotal, as those reusable target futures can be retrieved in the finetuning phase for fast adaptation.
We also find that the performance differences are more prominent on the medium-replay datasets.
This could be explained by the fact that medium-replay datasets consist of more diverse behaviors.
In such cases, controllable sampling plays an important role to distinguish the desired ones.

\paragraph{More ablation studies.}
We also compare {\ours} with its variants to examine how the inclusion of future latent variables affects performance.

\begin{figure}[t]
    \centering

    \begin{subfigure}[b]{0.49\linewidth}
         \centering
         \includegraphics[width=\linewidth]{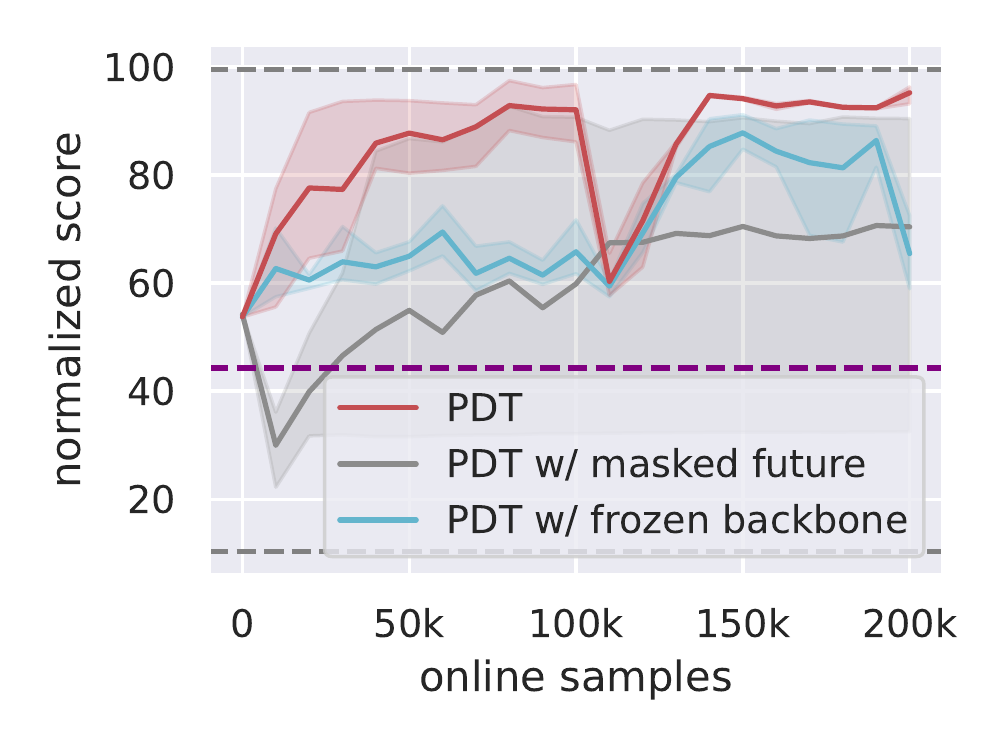}
         \caption{hopper-medium}
    \end{subfigure}
    \begin{subfigure}[b]{0.49\linewidth}
         \centering
         \includegraphics[width=\linewidth]{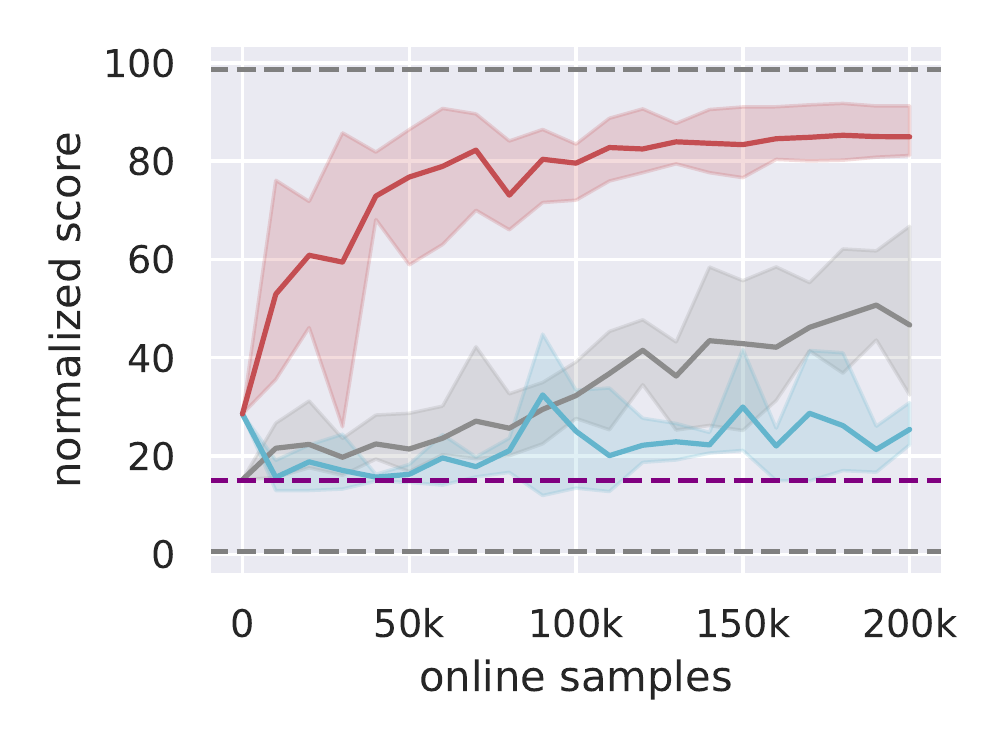}
         \caption{hopper-medium-replay}
     \end{subfigure}
    \caption{
    \textbf{Ablation results of {\ours} training.}
    We report finetuning results averaged over 3 random seeds.
    The shaded area shows the 95\% confidence interval.
    The \textcolor{mypurple}{purple} dashed line represents the averaged episodic return of offline trajectories, whereas the dashed lines in \textcolor{mygray}{gray} represent the min and max values.
    }
    \label{fig:freeze}
\end{figure}

To investigate how future conditioning contributes to pretraining performance, we compare {\ours} with a variant that masks out future embedding input for action prediction during pretraining.
Masking out future embeddings disables {\ours} to take actions based on the future and hence is helpful in ascertaining whether {\ours} benefits from future information or just memorizes offline behaviors.
We observe that pretraining with future embeddings masked leads to significantly degraded performance, as shown in Figure~\ref{fig:freeze}.

Next, we compare {\ours} with a variant that freezes all the parameters except for those of the return predictor during finetuning.
Figure~\ref{fig:freeze} shows that, by mining reward-maximizing actions out of diverse behaviors, the return prediction network plays an essential role so that {\ours} can continually refine its actions in the finetuning phase.
On the flip side, the results also demonstrate that solely relying on controllable sampling is insufficient for online finetuning.

\paragraph{Generalization.}
In the previous experiments we describe the performance of different pretraining algorithms on the standard D4RL datasets.
However, given that the offline trajectories used for pretraining are collected from agents solving the same tasks, it raises a question on how well do the pretrained models generalize to a variety of downstream tasks.
To examine this ability of {\ours}, we modify the reward functions (i.e., task specifications) of halfcheetah and walker2d in the finetuning phase and test whether the pretrained models can quickly adapt to new tasks.

Results are presented in Table~\ref{tab:generalization}.
After 200k online transitions of downstream task, {\ours} outperforms ODT by a large margin.
This reveals the advantages of unsupervised pretraining over supervised pretraining.
Since ODT learns from reward-labeled data, it tends to rely on target returns for decision making and hence struggles to improve when a new task is specified.
In contrast, {\ours} associates decisions to the task-agnostic future information, which can generalize well across different tasks.

\section{Conclusion}~\label{sec:conclusion}

In this paper, we present Pretrained Decision Transformer (PDT), an unsupervised 
pretraining algorithm for reinforcement learning (RL).
By learning to act based on the future,
{\ours} is able to extract rich prior knowledge from offline data.
This ability to leverage future information can be further exploited in the finetuning phase, as {\ours} associates each future possibility to its corresponding return and samples the one with the highest predicted return to make better decisions.
Experimental results demonstrate the effectiveness of {\ours} in comparison to a variety of competitive baselines.

One limitation of {\ours} is that it requires more training time and computational resources compared to DT and ODT.
This could result in practical challenges when the available resources are limited.
Besides, the objective in Equation~\ref{eq:future} creates a trade-off between diversity of the learned behaviors and behavior consistency.
Empirically we find that the optimal value is dataset-specific.
To improve future encoding, more advanced techniques such as VQ-VAE~\cite{van2017neural} could be applied.
For future work, we are interested in exploring how more expressive generative methods (e.g., diffusion models as policies~\cite{janner2022planning,wang2022diffusion}) can benefit {\ours}.

\begin{table}[t]
\centering
\begin{tabular}{l|ll}
\thickhline
task & ODT      & {\ours}      \\
\hline
halfcheetah-forward-jump & \textbf{87.27} {\tiny $\pm$ 14.41}
 & 83.80 {\tiny $\pm$ 2.28}
 \\
halfcheetah-jump         & -31.00 {\tiny $\pm$ 49.08}
& \textbf{70.39} {\tiny $\pm$ 16.56}
\\
walker2d-forward-jump    & 29.36 {\tiny $\pm$ 4.55}
& \textbf{45.31} {\tiny $\pm$ 36.81}
\\
walker2d-jump            & 15.81 {\tiny $\pm$ 14.75}
& \textbf{68.70} {\tiny $\pm$ 2.90}
\\
\hline
sum & 101.45 & \textbf{268.21} \\
\thickhline
\end{tabular}
\caption{
\textbf{Generalization performance.}
We run each instance for 200k online transitions and report the averaged normalized return over 3 random seeds.
See Appendix~\ref{appendix:generalization} for more details.}
\label{tab:generalization}
\end{table}


\section*{Acknowledgements}

The corresponding author Shuai Li is supported by National Key Research and Development Program of China (2022ZD0114804) and National Natural Science Foundation of China (92270201).







\bibliography{ref}
\bibliographystyle{icml2023}

\newpage
\appendix
\onecolumn

\section{Comparison to OPAL}\label{appendix:opal}

The learning objective of future embeddings in Equation~\ref{eq:future} bears resemblance to that of OPAL~\cite{ajay2021opal}.
Both {\ours} and OPAL encode state-action pairs into a latent space and use a latent-conditioned policy as the action decoder.
However, we want to clarify their differences as follows:

\begin{itemize}
    \item \textbf{Primitive behaviors vs future information.}
    Essentially, OPAL does not leverage future information for learning control. 
    For a given trajectory $\tau_{t: t+K-1}$, it extracts a latent vector and maximizes the conditional log-likelihood of actions in $\tau_{t: t+K-1}$ given the state and the latent vector. 
    Since the latent vector already contains information about the actions in $\tau_{t: t+K-1}$, OPAL aims to distill behaviors into temporally extended primitives, rather than to learn a future-conditioned policy. 
    In contrast, {\ours} learns a future encoder that embeds future trajectory $\tau_{t+K:t+2K-1}$ into latent space so that the latent-conditioned policy can leverage this privileged context to predict actions in $\tau_{t: t+K-1}$.

    \item \textbf{Single latent vs latent sequence.}
    OPAL encodes the whole trajectory of state-action pairs into a single latent vector, whereas {\ours} encodes the future trajectory into a sequence of latent vectors with causal masking.
    Similarly to the return-to-go sequence in DT~\cite{chen2021decision,zheng2022online}, latent sequences allow {\ours} to perceive more future information along the trajectory.

    \item \textbf{Regularization.} 
    OPAL does not regularize the latent distribution, whereas {\ours} regularizes by minimizing the KL divergence between the latent distribution and the standard Gaussian distribution. 
    There is no rule of thumb as to which is better.
    Empirically, we found that applying regularization is stable, probably because it can prevent {\ours} from over-reliance on the privileged future information (see Figure~\ref{fig:finetunekl}).
    However, we acknowledge that this design could be further improved, in directions like mixture of Gaussians regularization~\cite{tomczak2018vae}.
\end{itemize}

\section{Pseudocode for {\ours} Finetuning}~\label{appendix:pseudo}

\begin{algorithm}[ht]
\caption{Online Finetuning}
\label{algo:finetune}
\begin{algorithmic}[1]
\INPUT policy $\pi_\theta$, future prior $p_\theta$, future encoder $g_\theta$, return prediction network $f_\theta$, replay buffer $\mathcal{D}$, context length $K$, batch size $B$, number of online rollouts iteration $N$, training iteration $I$

\STATE Initialize replay buffer $\hat{\mathcal{D}}$ with online rollouts
\STATE Warmup return predictor $f_\theta$ with $\hat{\mathcal{D}}$
\FOR{$n = 1, \ldots, N$}
    \STATE $\hat{\tau} \gets$ rollout trajectory with $\pi_\theta$, $p_\theta$, $f_\theta$ using the controllable sampling mechanism in Section~\ref{sec:finetuning}
    \STATE $\hat{\mathcal{D}} \gets \{\hat{\mathcal{D}}\} \cup \{\hat{\tau}\}$ 

    \FOR{$i = 1, \ldots, I$}
        \STATE Sample $B$ trajectories from $\hat{\mathcal{D}}$ according to $p(\hat{\tau})=\hat{R}(\hat{\tau}) / \sum_{\hat{\tau}^\prime \in \hat{\mathcal{D}}}\hat{R}(\hat{\tau}^\prime)$
    
        \FOR{each sampled trajectory $\hat{\tau}$}
            \STATE $\hat{\tau}_{t:t+K-1} \gets$ a length-$K$ sub-trajectory uniformly sampled from $\hat{\tau}$
            \STATE $\hat{\tau}_{t+K:t+2K-1} \gets$ a length-$K$ future sub-trajectory
            \STATE Sample $z \sim g_{\theta}(\cdot | \tau_{t+K:t+2K-1})$
            \STATE Predict actions $a_{t^\prime} \sim \pi_\theta(\cdot \mid \tau_{t: t^\prime-1}, s_{t^\prime}, z)$, $\forall t^\prime = t, \ldots, t + K - 1$
            \STATE Predict returns $\hat{R}_{t^\prime} \sim f_\theta(\cdot \mid s_{t^\prime}, z)$, $\forall t^\prime = t, \ldots, t + K - 1$
            \STATE Calculate $\mathcal{L} = \mathcal{L}_\mathrm{BC} + \mathcal{L}_\mathrm{future} + \mathcal{L}_\mathrm{return}$ (Equation~\ref{eq:main_dt},\ref{eq:future},\ref{eq:return})
        \ENDFOR
        \STATE Update $\theta$ by gradient descent
    \ENDFOR
\ENDFOR
\end{algorithmic}
\end{algorithm}

\section{Experiment Details}~\label{appendix:details}

We conduct our experiments on a GPU cluster with 8 Nvidia 3090 graphic cards.
Pretraining {\ours} for 50k gradient steps on a single GPU typically takes 2-3 hours, whereas finetuning for 200k environment steps takes 6-8 hours.

\subsection{Hyperparameters}~\label{appendix:hyperparameter}

Our {\ours} implementation is based on the publicly available ODT codebase\footnote{\url{https://github.com/facebookresearch/online-dt}}.
We parameterize the future encoder $g_\theta$ using the same Transformer network as the policy.
The future prior network $p_\theta$ uses a pair of fully connected networks with 2 hidden layers and ReLU activation to obtain means and variances of future embedding.
The return predictor $f_\theta$ also uses a pair of fully connected networks with 2 hidden layers and ReLU activation to obtain a 1-dimensional Gaussian variable.

We use the LAMB optimizer~\cite{You2020Large} to jointly optimize the policy $\pi_\theta$, the future encoder $g_\theta$, the future prior $p_\theta$, and the return prediction network $f_\theta$.
The temperature parameter $\alpha$ is optimized by the Adam optimizer~\cite{kingma2014adam}.
Table~\ref{tab:hyperparameter} summaries the common hyperparameters for {\ours}.
For coefficient $\beta$, we search for each task in the range of \{1, 1e-2, 1e-3, 1e-4\} on the main experiments, and fix the choice for other experiments.
Table~\ref{tab:task_hyperparameter} lists the dataset-specific hyperparameters.

For ACL, we choose the most competitive unsupervised variant (i.e., bidir (T) + inputA/R (F) + finetune (T)) as reported in \citet{yang2021representation}.
We also take into account the recommended embedding dimension (512) and the pretraining window size (4) that are reported to work best for online RL finetuning.
ACL is pretrained for 200k gradient steps following the original paper.
For ODT, we use the reported hyperparameters.
As suggested in \citet{zheng2022online}, a long period of pretraining might hurt the exploration performance.
Therefore, we use the model checkpoint for each task once the pretrained model reaches the reported performance of offline pretraining.
For finetuning, all ACL, ODT, and {\ours} collect 10k online transitions to initialize the replay buffer at the beginning.

\begin{table}[ht]
    \centering
    \begin{tabular}{ll}
        \thickhline
        description & value \\
        \hline
        number of layers & 4 \\
        number of attention heads & 4 \\
        embedding dimension & 512 \\
        future latent dimension & 16 \\
        training context length & 20 \\
        evaluation context length & 5 \\
        positional embedding & no \\
        future sampling batch size & 256 \\
        return prediction warmup steps & 1500 \\
        dropout & 0.1 \\
        nonlinearity function & ReLU \\
        batch size & 256 \\
        learning rate &  0.0001 \\
        weight decay & 0.001 \\
        gradient norm clip & 0.25 \\
        learning rate warmup steps & $10^4$ \\
        target entropy & $-\text{dim}(\mathcal{A})$ \\
         \thickhline
    \end{tabular}
    \caption{Common hyperparameters that are used to train {\ours} in all the experiments.}
    \label{tab:hyperparameter}
\end{table}
\begin{table}[ht]
    \centering
\begin{tabular}{l|ccc}
\thickhline
                         dataset & pretraining updates & $\beta_\textrm{pretrain}$ & $\beta_\textrm{finetune}$ \\
\hline
hopper-medium             & 20000          & 1e-3       & 1e-4      \\
hopper-medium-replay      & 50000          & 1           & 1e-4      \\
walker2d-medium           & 20000          & 1e-3       & 1           \\
walker2d-medium-replay    & 50000          & 1           & 1e-4      \\
halfcheetah-medium        & 50000          & 1           & 1           \\
halfcheetah-medium-replay & 50000          & 1           & 1           \\
ant-medium                & 20000          & 1e-3       & 1           \\
ant-medium-replay         & 20000          & 1e-2        & 1e-2       \\
\thickhline
\end{tabular}
    \caption{
    Hyperparameters we use to train {\ours} for each dataset.}
    \label{tab:task_hyperparameter}
\end{table}

\subsection{Details about Rewardless-DT Baseline}\label{appendix:rewardless}

\begin{figure}[ht]
    \centering

    \begin{subfigure}[b]{0.245\linewidth}
         \centering
         \includegraphics[width=\linewidth]{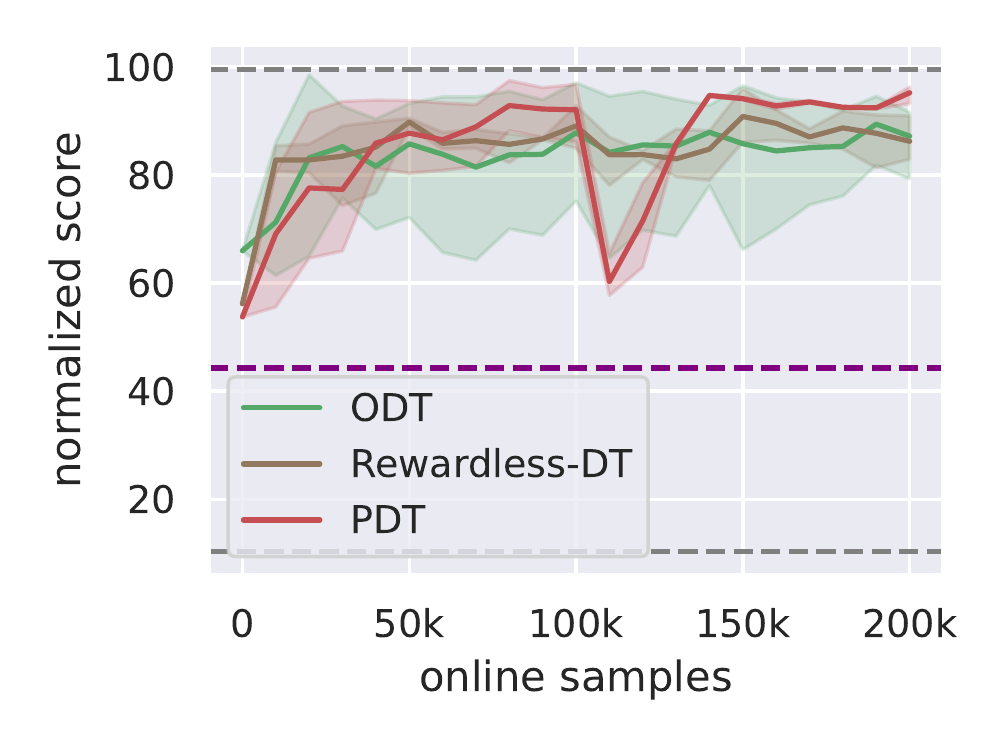}
         \caption{hopper-medium}
    \end{subfigure}
    \begin{subfigure}[b]{0.245\linewidth}
         \centering
         \includegraphics[width=\linewidth]{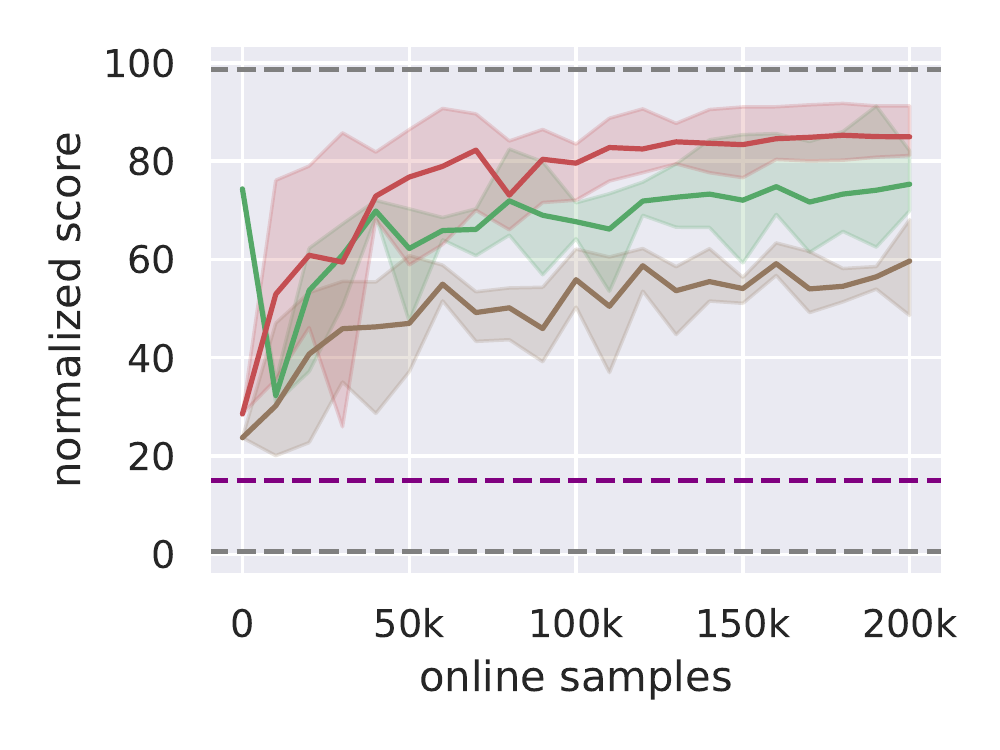}
         \caption{hopper-medium-replay}
     \end{subfigure}
    \begin{subfigure}[b]{0.245\linewidth}
         \centering
         \includegraphics[width=\linewidth]{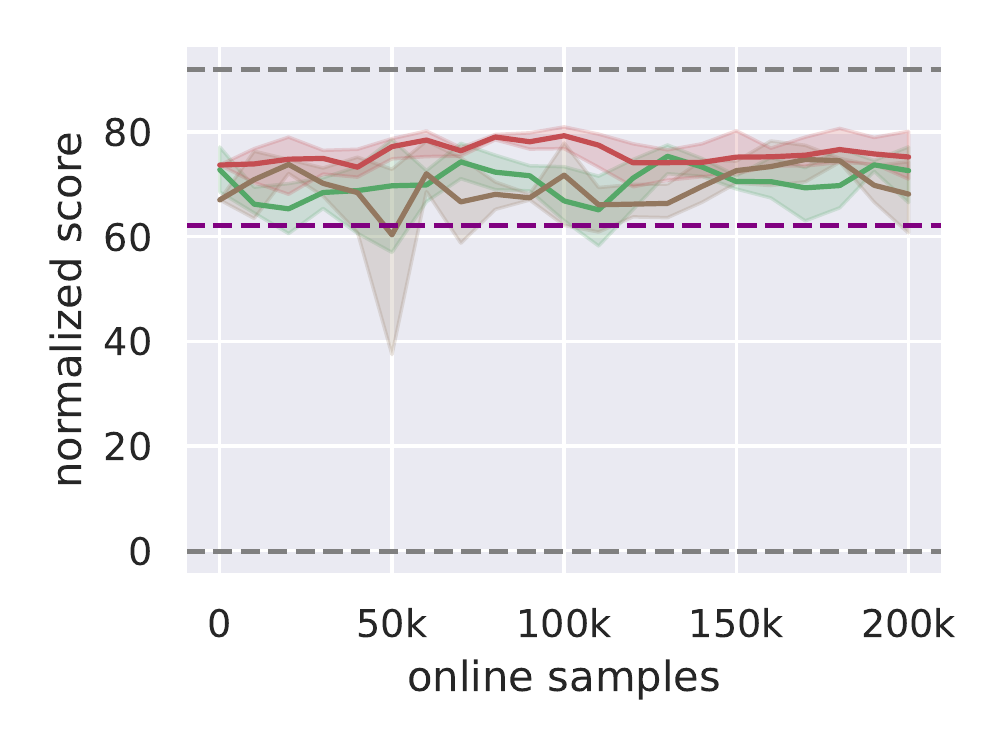}
         \caption{walker2d-medium}
    \end{subfigure}
    \begin{subfigure}[b]{0.245\linewidth}
         \centering
         \includegraphics[width=\linewidth]{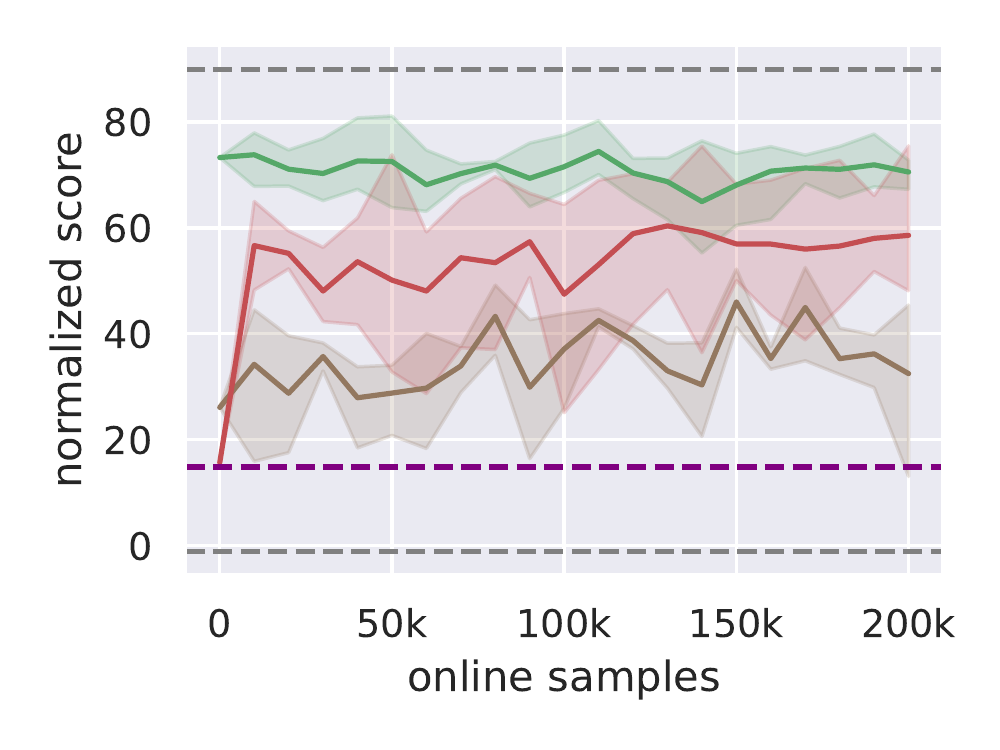}
         \caption{walker2d-medium-replay}
     \end{subfigure}
    \caption{
    \textbf{Comparison between {\ours} and Transformer-based baselines} on downstream finetuning performance.
    Each instance is evaluated over 10 episodes every 10k transitions.
    Results are averaged over 3 random seeds.
    The shaded area shows the 95\% confidence interval.
    The \textcolor{mypurple}{purple} dashed line represents the averaged episodic return of offline trajectories, whereas the dashed lines in \textcolor{mygray}{gray} represent the min and max values.
    See Appendix~\ref{appendix:rewardless} for more details about Rewardless-DT.
    }
    \label{fig:rewardless-dt}
\end{figure}

Rewardless-DT has the same network architecture as ODT.
The only difference is that, during pretraining, the return embeddings are masked to enable unsupervised learning.
We use the same training protocol and hyperparameters as those of ODT to train Rewardless-DT.
Figure~\ref{fig:rewardless-dt} compares the performance of Rewardless-DT, ODT, and {\ours} on four Gym MuJoCo tasks.

\subsection{Details about Future Conditioning Evaluation}~\label{appendix:future}

The analysis is done with an {\ours} agent pretrained on the walker2d-medium dataset, using the same hyperparameters reported in Appendix~\ref{appendix:hyperparameter}.
The pretrained {\ours} is evaluated on the walker2d task and conditioned on different future embeddings sampled from the pretrained future prior.
Each subfigure in Figure~\ref{fig:pretraining} shows the distributions of action on one dimension of the action space.
The description for each dimension can be found in the Gym documentation\footnote{\url{https://www.gymlibrary.dev/environments/mujoco/walker2d/}}.

\begin{figure}[ht]
    \centering

    \begin{subfigure}[b]{0.495\linewidth}
         \centering
         \includegraphics[width=\linewidth]{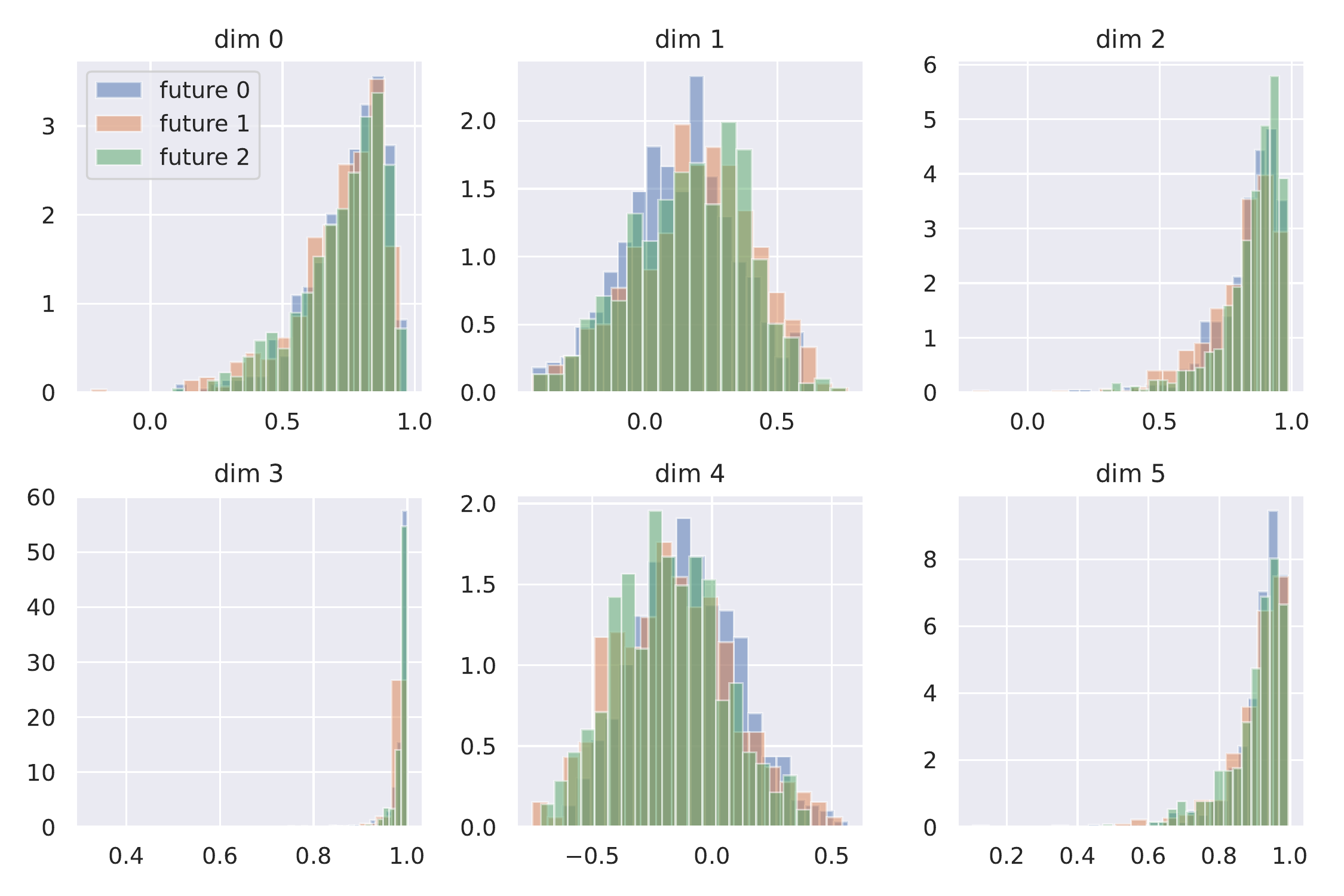}
         \caption{At timestep 500}
    \end{subfigure}
    \begin{subfigure}[b]{0.495\linewidth}
         \centering
         \includegraphics[width=\linewidth]{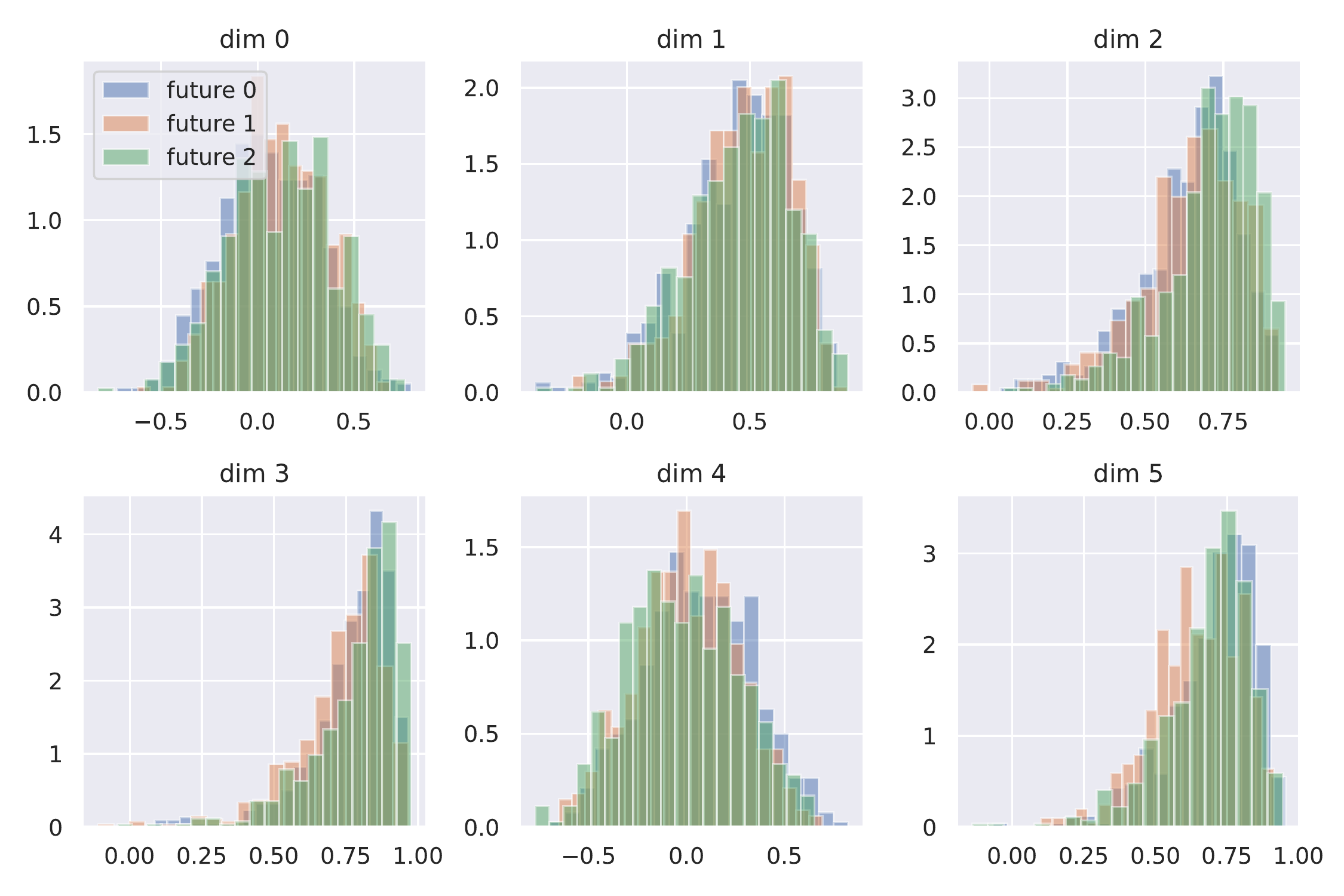}
         \caption{At timestep 1000}
    \end{subfigure}
    \caption{
    Comparison of the action distributions produced by {\ours} when conditioned on different future embeddings.
    We plot the histogram for each dimension
    of the action space.
    The result is taken from the state at timestep 500 (left) and timestep 1000 (right) of an episode when running {\ours} pretrained on the walker2d-medium dataset.
    See Appendix~\ref{appendix:future}.
    }
    \label{fig:appendix-pretrain}
\end{figure}

Figure~\ref{fig:appendix-pretrain} shows the action distributions of the same {\ours} model, at time step 500 and 1000 of the episode, respectively.

\subsection{Details about Controllable Sampling Evaluation}~\label{appendix:sampling}

The analysis is done with {\ours} agents pretrained on four D4RL datasets, using the same hyperparameters reported in Appendix~\ref{appendix:hyperparameter}.
But for finetuning, we modify the sampling procedure for online exploration and evaluation.
Instead of choosing the future embedding with the highest predicted return (100\%-percentile), we choose other percentiles.

\subsection{Details about Behavior Diversity Evaluation}\label{sec:behavior_diversity}

\begin{figure}[ht]
    \centering
    \includegraphics[width=0.6\linewidth]{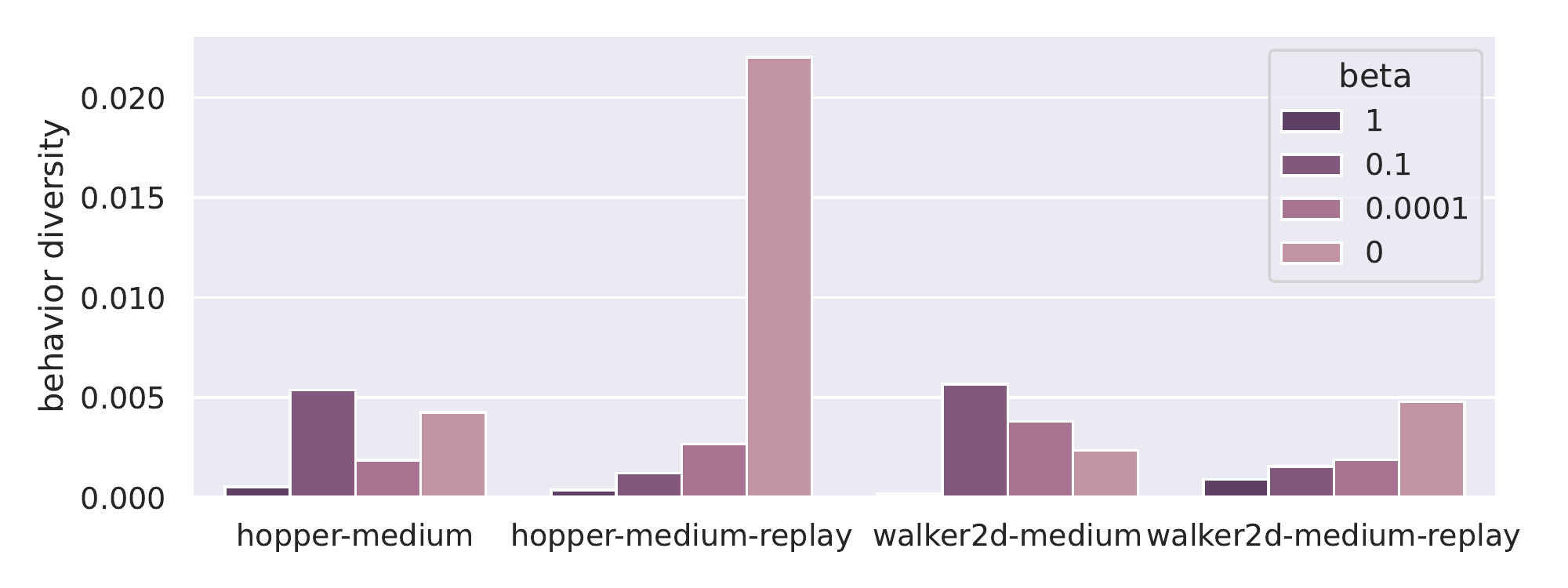}
    \caption{
    \textbf{Future regularization affects behavior diversity.}
    We compare the behavior diversity of {\ours} finetuned with different hyperparameter $\beta$.
    See Section~\ref{sec:behavior_diversity} for details.
}
    \label{fig:diversity}
\end{figure}

For each {\ours} policy, we evaluate its behavior diversity by how action distributions vary with different future embeddings to condition the policy.
The more dissimilar the action distributions are, the more diverse behaviors {\ours} can generate by sampling different future embeddings. 
Specifically, for each timestep, we sample 10 different future latent sequences from the future prior and obtain the corresponding action distributions $P_1, \ldots P_{10}$.
We measure their dissimilarity by average KL divergence: $D_{\text{KL}}\left(P_1, \ldots P_k\right)=\frac{1}{k(k-1)} \sum_{i, j=1}^k D_{\mathrm{KL}}\left(P_i \| P_j\right)$.
The dissimilarity is then averaged over all the timesteps of 10 episodes, resulting in a scalar value for each policy.
We use the distributions before applying the squashing function to calculate KL divergence. 
Results are reported in Figure~\ref{fig:diversity}.

\subsection{Details about Generalization Tasks}~\label{appendix:generalization}

We follow \citet{yu2021conservative} to construct four downstream tasks by changing the reward functions\footnote{The original reward functions can be found in \url{gymlibrary.dev}.}.
Specifically, we set the reward functions of task halfcheetah-forward-jump and halfcheetah-jump as $r(s, a) = -0.1 *\|a\|_2^2 + \text{vel}_x +15 * \text{pos}_z$ and $r(s, a) = -0.1 *\|a\|_2^2 + 15 * \text{pos}_z$.
For walker2d, we set the reward functions of task walker2d-forward-jump and walker2d-jump as $r(s, a) = -0.001 *\|a\|_2^2 + \text{vel}_x + 10 * \text{pos}_z$ and $r(s, a) = -0.001 *\|a\|_2^2 + 10 * \text{pos}_z$.
We use the same hyperparameters as in Appendix~\ref{appendix:hyperparameter} for downstream online RL.
We report scores normalized by computing $100 \times \frac{\text {score-random score}}{\text {best score-random score}}$.











\end{document}